\documentclass[manuscript,screen]{acmart}

\usepackage{enumitem}
\usepackage{rotating} 
\usepackage{wrapfig} 
\usepackage{float}
\usepackage{balance}
\renewcommand\footnotetextcopyrightpermission[1]{} 

\usepackage[tikz]{bclogo}

\usepackage{alltt}
\usepackage[skins]{tcolorbox}
\usepackage{multirow}
\usepackage{color}
\usepackage{colortbl} 
\definecolor{orange}{rgb}{1,0.647,0}
\definecolor{mypink}{HTML}{FFCCC9}
\AtBeginDocument{%
  \providecommand\BibTeX{{%
    \normalfont B\kern-0.5em{\scshape i\kern-0.25em b}\kern-0.8em\TeX}}}

\newenvironment{RQ}[1]%
{\noindent\begin{minipage}[c]{\linewidth}%
\begin{bclogo}[couleur=mypink!75,%
                arrondi=0.1,%
                logo=\bctrombone,%
                ombre=true]{{\small ~#1}}}%
{\end{bclogo}\end{minipage}\vspace{2mm}}
\AtBeginDocument{%
  \providecommand\BibTeX{{%
    \normalfont B\kern-0.5em{\scshape i\kern-0.25em b}\kern-0.8em\TeX}}}

\begin{document}

\title{Fair Enough:   Searching for Sufficient  Measures of Fairness}


\author{Suvodeep Majumder}
\email{smajumd3@ncsu.edu}
\affiliation{%
\institution{North Carolina State University}
\city{Raleigh}
\country{USA}}

\author{Joymallya Chakraborty}
\email{jchakra@ncsu.edu}
\affiliation{%
\institution{North Carolina State University}
\city{Raleigh}
\country{USA}}

\author{Gina R. Bai}
\email{rbai2@ncsu.edu}
\affiliation{%
\institution{North Carolina State University}
\city{Raleigh}
\country{USA}}

\author{Kathryn T. Stolee}
\email{ktstolee@ncsu.edu}
\affiliation{%
\institution{North Carolina State University}
\city{Raleigh}
\country{USA}}

\author{Tim Menzies}
\email{timm@ieee.org}
\affiliation{%
\institution{North Carolina State University}
\city{Raleigh}
\country{USA}}

\renewcommand{\shortauthors}{Majumder et al.}

\begin{abstract}
Testing machine learning software for ethical bias has become a pressing current concern.  In response, recent research has proposed a plethora of new fairness metrics, for example, the dozens of fairness metrics in the IBM AIF360 toolkit.  This raises the question: How can any fairness tool satisfy such a diverse range of goals? While we cannot completely simplify the task of fairness testing, we can certainly reduce the problem. This paper shows that many of those fairness metrics effectively  measure the same thing. Based on experiments using seven real-world datasets, we find that (a)~26 classification metrics can be clustered into seven groups, and (b)~four dataset metrics can be clustered into three groups. Further, each reduced set may actually predict  different things. Hence, it is no longer necessary (or even possible) to satisfy all  fairness metrics. In summary, to simplify the fairness testing problem, we recommend the following steps: (1)~determine what type of fairness is desirable (and we offer a handful of such types); then (2)~lookup those types in our clusters; then (3)~just test for one item per cluster. 

For the purpose of reproducibility,  our scripts and data are available at  {\bf \url{https://github.com/Repoanonymous/Fairness\_Metrics}}.
\end{abstract}




\maketitle

\section{Introduction}
\label{sec:intro}
A journal on software engineering methodologies needs to concern itself
not just with one applications, but also general methods that hold across multiple applications. Recently the authors faced a methodological issue where reviewers challenged the validity of the  metrics they  used to assess
that work. Prompted by that experience, we examined how we the current SE research community selects
metrics for   assessing the {\em fairness of algorithmic decision making}.

On reading the literature, we found a 
a general pattern: while the literature proposes a   propose a plethora
of metrics\footnote{E.g. the  Fairlearn~\cite{Fairlearn} tool 
 lists 16 metrics;
the Fairkit-learn tool~\cite{johnson2020fairkit} 
 comes  with  its own 16 metrics;    
IBM AIF360 toolkit~\cite{IBM}  offers 45 fairness metrics.},
we could not find  a principled argument (across a large space of known metrics) that it was necessary/unnecessary to report some metric X. This raises various methodological questions:

\begin{itemize}
    \item Should we reject papers that ``only'' use (e.g.) five metrics? Or should researchers always use dozens of metrics?
    \item When we use automatic tools to optimize for fairness, should we optimize for dozens of goals? 
     Or is optimizing for a  smaller set sufficient? 
\end{itemize}
To resolve these methodological concerns, we made the following conjecture.
Given, the large space of known
metrics (such as the 30 studied in this paper), perhaps
{\em many of these metrics are measuring the same thing}. As shown by the experiments of this paper, this is indeed the case,  since we can cluster these 30 metrics into around half a dozen. While our results pertain a particular domain, there is nothing in principle  stopping this methodology being applied to any domain where researchers keep proposing new metrics without first checking if the new metric is not just ``old wine in new bottles''

As to the specifics of our domain, this paper concerns itself with measures
of algorithmic fairness.
Increasingly, software is being used for critical decision-making processes, such as patient release from hospitals~\cite{7473150, Medical_Diagnosis}, credit card applications~\cite{10.1093/imaman/4.1.43}, hiring~\cite{AI_hiring}, and admissions~\cite{AI_admission}. According to guidelines from the  European Union~\cite{EU} and IEEE~\cite{IEEEethics}, a software cannot be used in real-life applications if it is found to be discriminatory toward an individual based on any sensitive attribute such as gender, race, or age.  Hence ``fairness testing'' is now an open and pressing problem in software engineering.

As shown in Table~\ref{tbl:fairness_def}, researchers  have proposed a plethora of  fairness metrics for measuring fairness, and that number is growing (e.g. see all
the metrics proposed in~\cite{Fairlearn,johnson2020fairkit,IBM}).  
Given that trend, is is somewhat strange to report that   researchers   in this area only use a  few metrics in their papers~\cite{pleiss2017fairness, 10.1145/3287560.3287563, kallus2018residual, wadsworth2018achieving, celis2018ranking, feldman2015certifying}. For example, in our literature review papers from the last three years, we see only a handful of papers (13 out of 60 to the best of our knowledge) using more than five fairness metrics to evaluate their method. This is surprising since all of them ignore more than half the known metrics. Is that wise?

The conjecture that is tested by this paper is that  {\em too many spurious metrics} which  {\em all measure very similar things}. If that were true, then it should be possible to simplify fairness assessment as follows: 
\begin{quote}
 {\em Run  metrics on real-world data. Find clusters of correlated metrics. Prune ``insensitive clusters\footnote{Note: Here, by ``insensitive'' clusters, we mean those where changes to the data do not change the fairness scores.}''. Only use one metric per surviving cluster.}
\end{quote}



This paper experiments with seven datasets and finds that (a)~26 classification fairness metrics can be clustered into just seven groups;  (b)~four dataset metrics can be clustered into three groups and that (c)~these clusters actually predict for different things. That is, it is no longer necessary (or even possible) to satisfy all these fairness metrics. Hence, to simplify fairness testing, we recommend (a)~determining what type of fairness is desirable (and we offer a handful of such types); then (b)~looking up those types in our clusters; then (c)~testing for one item per cluster.

This paper is  structured  around these research questions.

\textbf{RQ1:} {\em Do current fairness metrics agree with each other? } Our experiments show that current fairness metrics often disagree, markedly.

\textbf{RQ2:} {\em Can we group (cluster) fairness metrics based on similarity? } We find   sets of similar metrics using agglomerative clustering~\cite{AgglomerativeClustering}.

\textbf{RQ3:} {\em  Are   some fairness metrics more sensitive  to change than others?} While most are sensitive, some are not.

\textbf{RQ4:} {\em Can we achieve fairness based on all the metrics at the same time? } It is  challenging to do so since some of them are competing goals and some are contradictory based on definitions.


\begin{table*}[]
\caption{Mathematical definitions of the classification and dataset metrics used in this research. Definitions are collected from IBM AIF360~\cite{IBM}, Fairkit-learn~\cite{johnson2020fairkit} \& Fairlearn~\cite{Fairlearn}. For definitions of the terms used here, see Table~\ref{Confusion_Matrix}.}
\resizebox{\textwidth}{!}{%
\begin{tabular}{|c|l|l|c|c|c|c|}
\hline
\rowcolor[HTML]{C0C0C0} 
\begin{tabular}[c]{@{}c@{}}Metric Id\\ (MID)\end{tabular} & \multicolumn{1}{c|}{Metric Name}                                                             &  \multicolumn{1}{c|}{Description} & Ideal Value & AIF360 & Fairkit & Fairlearn \\ \hline
\rowcolor[HTML]{C0C0C0}  \multicolumn{7}{|c|}{{Classification Metrics}}                                                                   \\ \hline
C0          & true\_positive\_rate\_difference                                                             & $TPR_{D=\mathit{unprivileged}} - TPR_{D=\mathit{privileged}}$            & 0     & \checkmark & \checkmark  & \checkmark     \\ \hline
C1          & false\_positive\_rate\_difference                                                            & $FPR_{D=\mathit{unprivileged}} - FPR_{D=\mathit{privileged}}$             & 0    & \checkmark & \checkmark &  \checkmark      \\ \hline
C2          & false\_negative\_rate\_difference                                                            & $FNR_{D=\mathit{unprivileged}} - FNR_{D=\mathit{privileged}}$            & 0    & \checkmark & \checkmark & \checkmark       \\ \hline
C3          & false\_omission\_rate\_difference                                                            & $FOR_{D=\mathit{unprivileged}} - FOR_{D=\mathit{privileged}}$            &  0   & \checkmark & \checkmark &         \\ \hline
C4          & false\_discovery\_rate\_difference                                                           & $FDR_{D=\mathit{unprivileged}} - FDR_{D=\mathit{privileged}}$             & 0   & \checkmark & \checkmark &         \\ \hline
C5          & false\_positive\_rate\_ratio                                                                 & $FPR_{D=\mathit{unprivileged}} / FPR_{D=\mathit{privileged}}$            & 1   & \checkmark & \checkmark & \checkmark        \\ \hline
C6          & false\_negative\_rate\_ratio                                                                 & $FNR_{D=\mathit{unprivileged}} / FNR_{D=\mathit{privileged}}$            & 1  & \checkmark & \checkmark & \checkmark         \\ \hline
C7          & false\_omission\_rate\_ratio                                                                 & $FOR_{D=\mathit{unprivileged}} / FOR_{D=\mathit{privileged}}$            & 1  & \checkmark & \checkmark &          \\ \hline
C8          & false\_discovery\_rate\_ratio                                                                & $FDR_{D=\mathit{unprivileged}} / FDR_{D=\mathit{privileged}}$            & 1   & \checkmark & \checkmark &         \\ \hline
C9          & average\_odds\_difference                                                                    & \begin{tabular}[c]{@{}l@{}}$\frac{1}{2}(\mathit{false\_positive\_rate\_difference}$ \\ + $\mathit{true\_positive\_rate\_difference})$\end{tabular}            & 0  & \checkmark & \checkmark &          \\ \hline
C10         & average\_abs\_odds\_difference                                                               & \begin{tabular}[c]{@{}l@{}}$\frac{1}{2}(|\mathit{false\_positive\_rate\_difference}|$ \\ + $|\mathit{true\_positive\_rate\_difference}|)$\end{tabular}              & 0   & \checkmark & \checkmark &         \\ \hline
C11         & error\_rate\_difference                                                                      & $ERR_{D=\mathit{unprivileged}} - ERR_{D=\mathit{privileged}}$            & 0     & \checkmark & \checkmark &       \\ \hline
C12         & error\_rate\_ratio                                                                           &  $ERR_{D=\mathit{unprivileged}} / ERR_{D=\mathit{privileged}}$            & 1   & \checkmark & \checkmark &         \\ \hline
C13         & selection\_rate                                                                              & $Pr(\hat{Y} = favorable)$            & 0   & \checkmark & \checkmark &        \\ \hline
C14         & disparate\_impact                                                                            & $Pr(\hat{Y} = 1|D=\mathit{unprivileged})/Pr(\hat{Y} = 1|D=\mathit{privileged})$            & 1   & \checkmark & \checkmark & \checkmark        \\ \hline
C15         & statistical\_parity\_difference                                                              & $Pr(\hat{Y} = 1|D=\mathit{unprivileged}) - Pr(\hat{Y} = 1|D=\mathit{privileged})$            & 0  & \checkmark & \checkmark & \checkmark         \\ \hline
C16         & generalized\_entropy\_index                                                              & $\frac{1}{n\alpha(\alpha-1)}\sum_{i=1}^{n}[(\frac{b_{i}}{\mu})^\alpha - 1] $ where $b_{i} = \hat{y_{i}} - y_{i} + 1$             & 0  & \checkmark & &          \\ \hline
C17         & \begin{tabular}[c]{@{}l@{}}between\_all\_groups\_generalized\\ \_entropy\_index\end{tabular} & generalized\_entropy\_index between all groups            & 0  & \checkmark & &          \\ \hline
C18         & \begin{tabular}[c]{@{}l@{}}between\_group\_generalized\\ \_entropy\_index\end{tabular}       & \begin{tabular}[c]{@{}l@{}}generalized\_entropy\_index \\ between privileged and unprivileged groups\end{tabular}           & 0  & \checkmark & &          \\ \hline
C19         & theil\_index                                                                                 & $\frac{1}{n}\sum_{i=1}^{n}\frac{b_{i}}{\mu}\ln{\frac{b_{i}}{\mu}}$            & 0   & \checkmark & &         \\ \hline
C20         & coefficient\_of\_variation                                                                   & $2*\sqrt{\mathit{generalized\_entropy\_index}}$            & 0  & \checkmark & &          \\ \hline
C21         & between\_group\_theil\_index                                                                 & theil\_index between privileged and unprivileged groups            & 0    & \checkmark & &        \\ \hline
C22         & \begin{tabular}[c]{@{}l@{}}between\_group\_coefficient\\ \_of\_variation\end{tabular}        & coefficient\_of\_variation privileged and unprivileged groups            & 0   & \checkmark & &         \\ \hline
C23         & \begin{tabular}[c]{@{}l@{}}between\_all\_groups\_theil\\ \_index\end{tabular}                & theil\_index between all groups            & 0     & \checkmark & &       \\ \hline
C24         & \begin{tabular}[c]{@{}l@{}}between\_all\_groups\_coefficient\\ \_of\_variation\end{tabular}  & coefficient\_of\_variation between all groups           & 0  & \checkmark & &          \\ \hline
C25         & \begin{tabular}[c]{@{}l@{}}differential\_fairness\_bias\\ \_amplification\end{tabular}       & Smoothed EDF between the classifier and the original dataset              & 0   & \checkmark & &         \\ \hline
\rowcolor[HTML]{C0C0C0}  \multicolumn{7}{|c|}{Dataset Metrics}                                                                                                 \\ \hline
D0          & consistency                                                                                  &  $1 - \frac{1}{n*n\_neighbors}\sum_{i-1}^{n}|\hat{y_i} - \sum_{j\in{N_{n_neighbors(x_i)}}}\hat{j_y}|$          & 1 & \checkmark & &           \\ \hline
D1          & \begin{tabular}[c]{@{}l@{}}smoothed\_empirical\\ \_differential\_fairness\end{tabular}                                                  &  Smoothed EDF           & 0   & \checkmark & &         \\ \hline
D2          & mean\_difference                                                                             & $Pr(\hat{Y} = 1|D=\mathit{unprivileged}) - Pr(\hat{Y} = 1|D=\mathit{privileged})$            & 0  & \checkmark & &          \\ \hline
D3          & disparate\_impact                                                                            & $Pr(\hat{Y} = 1|D=\mathit{unprivileged})/Pr(\hat{Y} = 1|D=\mathit{privileged})$           & 1    & \checkmark & &        \\ \hline
\end{tabular}}
\label{tbl:fairness_def}
\end{table*}

In terms of research contributions,
this study is important since the art of software fairness testing is evolving rapidly. Studies like the one are essential to documenting what methods are ``best'' (as opposed to  those that might distract from   core issues). Accordingly:
\begin{itemize}
\item
   This paper proposes
a novel  metric assessment tactic
that can clarify and  simplify  future research
reports in this field (run metrics on real-world data; find clusters of correlated metrics; prune
``insensitive clusters1''; only use one metric per surviving cluster). 

\item
This paper tests that tactic  in an {\em extensive case study} applying 30 fairness metrics and grouped them into clusters  (RQ1 \& RQ2).  We say this study is extensive since
 it is far more detailed than prior work.
 All our empirical results were repeated 25 times. 
Our study explores multiple bias mitigation algorithms on seven  datasets (than prior work~\cite{grari2019fair, celis2019improved, Chakraborty2021BiasIM, Chakraborty_2020, cesaro2019measuring} was tested on far fewer metrics and far fewer datasets).
\item
To the best of our knowledge,
this study is the first one to perform such a {\em sensitivity meta-analysis} of fairness testing and to  warn that some metrics are unresponsive to data changes (RQ3). 
\item
This study also presents a {\em meta-analysis of metrics ability} to achieve fairness after application of bias mitigation technique (RQ4).  
\item
 In order to support replication and reproduction of our results, all  our datasets and scripts are publicly available at   https://github.com/Repoanonymous/Fairness\_Metrics.
\end{itemize}
\newpage

\subsection{Preliminaries}
Before beginning, we digress to make four  points. 

{\em Firstly}, mitigating the untoward effects of AI is a much broader problem than just exploring bias in algorithmic decision making (as done in this paper). The general problem of fairness is that influential groups in our society might mandate systems that (deliberately or unintentionally) disadvantage sub-groups within that society. An algorithm might satisfy all the metrics of Table~\ref{tbl:fairness_def} and still perpetuate social inequities. For example:

\begin{itemize}
    \item Its license feeds might be so expensive that only a small monitory of organizations can boast they are ``fair''; 
    \item The skills required to use a model's API might be so elaborate that only an elite group of programmers can use it even if the model is fair.
\end{itemize}

More generally,  Gebru et al.~\cite{gebru21, buolamwini2018gender} argues that inequities arise from the core incentives that drive the  organizations building an AI model, e.g., tools funded by the Defence Department have a tendency to support damage to property or life. She argues that ``There needs to be regulation that specifically says that corporations need to show that their technologies are not harmful before deploying them''. In terms of her work, this paper addresses the technical issue of how to measure ``harm''. As we show in Table~\ref{tbl:fairness_def} there are dozens of ways we might call software ``biased'' (and, hence, harmful). But we can also show is that many of those measures are relatively uninformative. Hence, if some organization wishes to follow the recommendations of Gebru et al., then with the methods of this paper, they can make their case of ``harmless'' via a smaller and simpler report. 

{\em Secondly}, Table~\ref{tbl:fairness_def} lists dozens of metrics currently seen in the  SE fairness testing literature. This paper makes an {\em empirical argument} that this list is too long since many of these metrics offer similar conclusions. One alternative to our {\em empirical argument}  is an {\em analytical argument} that  metric X  (e.g.) is equivalent to metric Y. Later in this paper (see \S\ref{sec:cluster_analysis}), we make the case that to reduce the space of metrics to be explored, that kind of  analytical argument may actually be misleading.

{\em Thirdly}, to be clear, while we can reduce dozens of metrics down to ten, there will still be issues of how to trade-off within this reduced set.  That said, we assert our work is valuable since debating the merits of, say, ten  metrics is a far more straightforward task than trying to resolve all the conflicts between 30. Further, and  more importantly, our methods could be used as a litmus test to prune away  spurious new  metrics that merely report old ideas but in a different way.

{\em Fourthly}, even after our mitigation algorithms, some fairness metrics still can contradict each other regarding the presence of bias. Hence, in  \S\ref{what_to_do}, we offer an extensive discussion on what to do in that situation.

\section{Background}

\subsection{The Problem of Algorithmic Fairness}

As software developers, we cannot turn a blind eye to the detrimental social effects of our software.  While no single paper can hope to fix all social inequities, this paper shows how to reduce the complexity involved in assessing one particular kind of unfairness (algorithmic decision making bias).
There is much evidence of machine learning (ML) software showing biased behavior. For example, language processing tools are more accurate on English written by Anglo-Saxons  than written by people of other races~\cite{blodgett2017racial}. An Amazon hiring tool was found to be biased against women~\cite{Amazon_Recruit}. YouTube makes more mistakes while generating closed captions for videos with female voices than males~\cite{tatman-2017-gender,Koenecke7684}. A popular risk-score predicting algorithm was found to be heavily biased against African Americans showing a higher error rate while predicting future criminals~\cite{propublica}. Gender bias is also prevalent in Google~\cite{Caliskan183} and Bing~\cite{johnson2020fairkit} translators. 

Due to so many undesirable events,  academic researchers and big industries have started giving immense importance to ML software fairness. Microsoft has launched ethical principles of AI where ``fairness'' has been given the topmost priority~\cite{MicrosoftEthics}. IBM has built a toolkit called AI Fairness 360~\cite{AIF360} containing the most noted works in the fairness domain.  In recent years, the software engineering research community has also started exploring this topic actively.  ICSE'18 held a special workshop for ``software fairness''~\cite{FAIRWARE}. ASE'19 held  another workshop called EXPLAIN, where fairness and explainability of ML models were discussed~\cite{EXPLAIN}. Johnson et al.  have created a public GitHub repository for data scientists to evaluate ML models based on quality and fairness metrics simultaneously~\cite{johnson2020fairkit}.

As to technology developed to detect and fix these issues of fairness, we can see three groups:  {\em fairness testing,  model-based mitigation, and fairness metrics, }.
    
\textbf{Fairness Testing}: The idea here is to generating discriminatory test cases and finding whether the model shows discrimination or not. The first work on this was THEMIS, done by Galhotra et al.~\cite{Galhotra_2017}. THEMIS generates test cases by randomly perturbing attributes. AEQUITAS~\cite{Udeshi_2018} improves the way of test case generation to become more efficient. Aggarwal et al. combined local explanation and symbolic execution to generate a better black-box testing strategy~\cite{Aggarwal:2019:BBF:3338906.3338937}.
    
\textbf{Model Bias Mitigation}: There are three techniques used to remove bias from model behavior. The first one is ``pre-processing'' where before model training, bias is removed from training data. Some popular prior work includes  optimized pre-processing~\cite{NIPS2017_6988}, Fair-SMOTE~\cite{Chakraborty2021BiasIM} and reweighing~\cite{Kamiran2012}. The second one is ``in- processing'' where after model training, the trained model is optimized for fairness. Popular prior work includes prejudice remover regularizer~\cite{10.1007/978-3-642-33486-3_3} and meta fair classifier~\cite{celis2020classification}. The last one is ``post-processing'' where while making predictions, model output is changed to remove discrimination. Some noted works  include reject option classification~\cite{Kamiran:2018:ERO:3165328.3165686} and  calibration~\cite{pleiss2017fairness}. There is some work that combines two or more of these techniques, such as Fairway~\cite{Chakraborty_2020}, a combination of ``pre-processing'' and ``in-processing''. 

While the fairness testing and model bias mitigation are important areas, we note that {\em before} we can declare success in those two areas, we {\em first} need some way to measure that success.

Accordingly, this paper focuses on the third area called:

\textbf{Fairness Metrics}:  Early work in this area was done by Verma et al.~\cite{10.1145/3194770.3194776} who divided 20 fairness metrics into five groups based on the theoretical definitions. Hinnefeld et al.  made a comparative analysis of four fairness metrics~\cite{hinnefeld2018evaluating}. Wang et al. did a user study to find a relation between fairness metrics and human judgments~\cite{wang2019empirical}. There are also some papers coming from industry on the topic. LinkedIn has created a toolkit called LiFT for scalable computation of fairness metrics as part of large ML systems~\cite{Vasudevan_2020}. Recently, Amazon internally published an empirical study based on  18 fairness metrics~\cite{Amazon_AWS}.

While all that research is certainly insightful, in some sense that work has been too successful. As mentioned in the introduction, the above work has now generated a  plethora of metrics. Hence, for the rest of this paper, we check if we can simplify the current space of metrics.

\begin{table*}[]
\caption{Details of the datasets used in this research.}
\label{datasets}
\small
\begin{tabular}{|c|c|c|c|c|c|c|}
\hline
\rowcolor[HTML]{C0C0C0} 
\textbf{Dataset}                                               & \textbf{\#Rows} & \textbf{\#Features} & \multicolumn{2}{c|}{\cellcolor[HTML]{C0C0C0}\textbf{Protected Attribute}}                                                                   & \multicolumn{2}{c|}{\cellcolor[HTML]{C0C0C0}\textbf{Class Label}} \\ \hline
\rowcolor[HTML]{C0C0C0} 
\textbf{}                                                      & \textbf{}       & \textbf{}           & \textbf{Privileged}                                               & \textbf{Unprivileged}                                                   & \textbf{Favorable}              & \textbf{Unfavorable}            \\ \hline
\begin{tabular}[c]{@{}c@{}}Adult Census\\ Income~\cite{ADULT}\end{tabular}  & 48,842          & 14                  & \begin{tabular}[c]{@{}c@{}}Sex-Male\\ Race-White\end{tabular}     & \begin{tabular}[c]{@{}c@{}}Sex-Female\\ Race-Non-white\end{tabular}     & High Income                     & Low Income                      \\ \hline
Compas~\cite{COMPAS}                                                        & 7,214           & 28                  & \begin{tabular}[c]{@{}c@{}}Sex-Male\\ Race-Caucasian\end{tabular} & \begin{tabular}[c]{@{}c@{}}Sex-Female\\ Race-Not Caucasian\end{tabular} & Did not reoffend                & Reoffended                      \\ \hline
German Credit~\cite{GERMAN}                                                 & 1,000           & 20                  & Sex-Male                                                          & Sex-Female                                                              & Good Credit                     & Bad Credit                      \\ \hline
Heart Health~\cite{HEART}                                                  & 297             & 14                  & Age-Young                                                         & Age-Old                                                                 & Not Disease                     & Disease                         \\ \hline
Bank Marketing~\cite{BANK}                                                & 45,211          & 16                  & Age-Old                                                           & Age-Young                                                               & Term Deposit - Yes              & Term Deposit - No               \\ \hline
\begin{tabular}[c]{@{}c@{}}Student \\ Performance~\cite{STUDENT}\end{tabular} & 1,044           & 33                  & Sex-Male                                                          & Sex-Female                                                              & Good Grade                      & Bad Grade                       \\ \hline
Titanic ML~\cite{TITANIC}                                                   & 891             & 10                  & Sex-Male                                                          & Sex-Female                                                              & Survived                        & Not Survived                    \\ \hline
\end{tabular}
\end{table*}

\subsection{Metrics Used in this Study}
\label{sec:Terminology}
In our work, we collected all the metric definitions from the IBM AI Fairness 360 GitHub repository. Table~\ref{tbl:fairness_def} lists the metrics studied in this paper. The \textit{Fairkit} and \textit{Fairlearn} columns in Table~\ref{tbl:fairness_def} show the metrics that are common among the IBM AIF360 metrics and metrics from Fairkit~\cite{johnson2020fairkit} (16 out of 16 available metrics) and Fairlearn~\cite{Fairlearn} (7 out of 16 metrics) toolkit. 

Before explaining fairness metrics, we need to understand some terminology. Table~\ref{datasets} contains seven binary classification datasets. The binary outcomes are  \textit{favorable} if it gives an advantage to the receiver (i.e., being hired for a job, getting credit card approved). Each of these datasets has at least one \textit{protected attribute} that divides the population into two groups (\textit{privileged \& unprivileged}) that have differences in terms of benefits received. ``sex'', ``race'', ``age'' are examples of protected attributes. The goal of group fairness is, based on the protected attribute, privileged and unprivileged groups will be treated similarly. While individual fairness tries to provide similar outcomes to similar individuals.

A \textit{fairness metric} is a quantification of unwanted bias in training data or models. Table~\ref{tbl:fairness_def} shows a sample of such metrics. When selecting these particular metrics, we skipped over:
\begin{itemize}
    \item Metrics for which we could not access precise definitions and implementations in IBM AIF360 toolkit~\cite{IBM};
    \item Metrics for which we could not find publications to use as baselines in this paper.
\end{itemize}

These two selection rules resulted in the 30 metrics of  Table~\ref{tbl:fairness_def}, which divide as follows:

\textbf{Classification Metrics:} These measure fairness based on classification results and are labeled in Table~\ref{tbl:fairness_def} using a \textit{Metric Id} beginning with \textit{C}. Two inputs are needed to measure this: the first one is original dataset with true labels and the second one is predicted dataset. In case of binary classification, classification metrics can be calculated from confusion matrix. Table~\ref{Confusion_Matrix} shows a combined confusion matrix where every cell is divided based on the protected attribute. 

\textbf{Dataset Metrics:} While classification metrics relate to predictions made from models, {\em dataset metrics} discuss learner-independent properties of the data. These are labeled in Table~\ref{tbl:fairness_def} using a \textit{Metric Id} beginning with \textit{D}. Only one input is needed to compute this: original dataset or transformed (by some bias mitigation algorithm) dataset. It can be applied for both \textit{group and individual fairness}.       

\textbf{Distortion Metrics:}
For completeness, we note that AIF360  includes  a third set of metrics called  {\em distortion metrics}. While these metrics are not seen extensively in the current literature, they would be a worthy target for future research.

In Table~\ref{tbl:fairness_def}, each metric has an {\em ideal value} representing the best-case scenario. This means at ideal value according to the metric privileged and unprivileged groups are treated equally. For most of the metrics, the ideal value is zero, while in some cases where the metric is a ratio, the ideal value is one. If the ideal value for a metric is zero, a positive value denotes an advantage for the unprivileged group, while a negative value  denotes an advantage for the privileged group. On the other hand, if the ideal value for a metric is one, a value < one denotes an advantage for the privileged group and > one denotes an advantage for the unprivileged group. 

To use these metrics, some threshold must be applied to report ``fair'' or ``unfair'';

\begin{itemize}
    \item For metrics with ideal value 0: the IBM AIF360 toolkit~\cite{IBM} uses the following definition of ``fair'': ranges between  -0.1 to 0.1 as ``fair'' (so ``unfair'' means values outside that range).
    \item For metrics with ideal value 1: the IBM AIF360 toolkit~\cite{IBM} uses the following definition of ``fair'': ranges between  0.8 to 1.2 as ``fair'' (so ``unfair'' means values outside that range).
\end{itemize}

\begin{table}[!t]
\caption{Mathematical definition of various confusion matrix based rates. These are used to calculate fairness metrics defined in Table \ref{tbl:fairness_def}.}.
\label{Confusion_Matrix}
\small
\begin{tabular}{|c|c|c|}
\hline
\rowcolor[HTML]{C0C0C0} 
                                                             & Actual Positive                                                              & Actual Negative                                                              \\ \hline
\begin{tabular}[c]{@{}l@{}}Predicted\\ Positive\end{tabular} & \begin{tabular}[c]{@{}l@{}}TP\\ PPV = TP/(TP+FP)\\ TPR = TP/(TP+FN)\end{tabular} & \begin{tabular}[c]{@{}l@{}}FP\\ FDR = FP/(TP+FP)\\ FPR = FP/(FP+TN)\end{tabular} \\ \hline
\begin{tabular}[c]{@{}l@{}}Predicted\\ Negative\end{tabular} & \begin{tabular}[c]{@{}l@{}}FN\\ FOR = FN/(TN+FN)\\ FNR = FN/(TP+FN)\end{tabular} & \begin{tabular}[c]{@{}l@{}}TN\\ NPV = TN/(TN+FN)\\ TNR=TN/(TN+FP)\end{tabular}   \\ \hline
\end{tabular}
\end{table}

\section{Methodology}
\label{sec:Methodology}
\subsection{Models}
\label{subsec:models}
This paper analyzes  the 30 fairness metrics in Table~\ref{tbl:fairness_def} using  the seven  datasets described in Table~\ref{datasets}.  In that work, we use  one baseline model and two models tuned by pre-processing and in-processing algorithms to compare with: 

\begin{itemize}
    \item \textbf{Baseline:} We used a logistic regression model for creating baseline results. Logistic regression is widely used in the fairness domain as baseline model~\cite{Chakraborty_2020,10.1007/978-3-642-33486-3_3,calmon2017optimized,chakraborty2019software,9286091}. We used  scikit-learn implementation with `l2' regularization (which helps to prevent over-fitting), `lbfgs' solver (which is a quasi-Newton optimization algorithm), and maximum iteration of 1000. 
    \item \textbf{Reweighing:} A widely used~\cite{cesaro2019measuring, agrawal2020debiasing, bellamy2019ai, sattigeri2019fairness, jones2020metrics} pre-processing method proposed by Kamiran et al.~\cite{Kamiran2012}. Here, before model training, examples in each group, and label are given different weights to ensure fairness. 
    \item \textbf{Meta Fair Classifier:} An in-processing method proposed by Celis et al.~\cite{celis2020classification}, which is a widely used meta algorithm~\cite{celis2019improved, grari2019fair, bera2019fair, padh2020addressing}. The optimization algorithm is developed to improve 11 fairness metrics with minimal loss in accuracy.
\end{itemize}

The last two bias mitigation algorithm implementations are taken from IBM AIF360~\cite{aif360-oct-2018}.

\subsection{Agglomerative Clustering}
\label{subsec:agglo}

Our metrics selection strategy,  requires a clustering algorithm. Two class of such clustering algorithms are  (a)~partitioning clustering and (b)~hierarchical clustering.  Here we are grouping fairness metrics based on similarity, not on distance, and we have no prior idea about the number of clusters. Thus, in this case, the ideal choice is \textit{hierarchical clustering}. Agglomerative clustering~\cite{AgglomerativeClustering} is a hierarchical bottom-up clustering approach that is widely used in the ML community~\cite{dickinson2001finding, zhang2016cross, rodrigues2008hierarchical, shirkhorshidi2015comparison, do2008clustering, d2005does, lin2016log, fokaefs2011jdeodorant, bavota2014recommending}.  In this approach, the closest pairs of items are grouped together. These closest of these groups are then grouped into a higher-level group. This repeats until everything falls into one group. We used the average pairwise dissimilarity between objects in two different clusters as linkage method between groups. This process creates a dendrogram, a hierarchical structure of the groups/clusters obtained by between-cluster distance or dissimilarity. From this tree of groupings, we use the within-cluster similarity from the dendrogram,  look for the largest distance that we can travel vertically without crossing any horizontal line~\cite{thorndike1953belongs, stanford, dann2017reconstructing}, and extract the clusters at the largest change in dissimilarity (which is similar to SSE - Sum of Squared Error).

\subsection{Spearman Rank Correlation}
\label{subsec:corr}

\begin{figure}[!t]
\centering
    \includegraphics[width=0.45\linewidth]{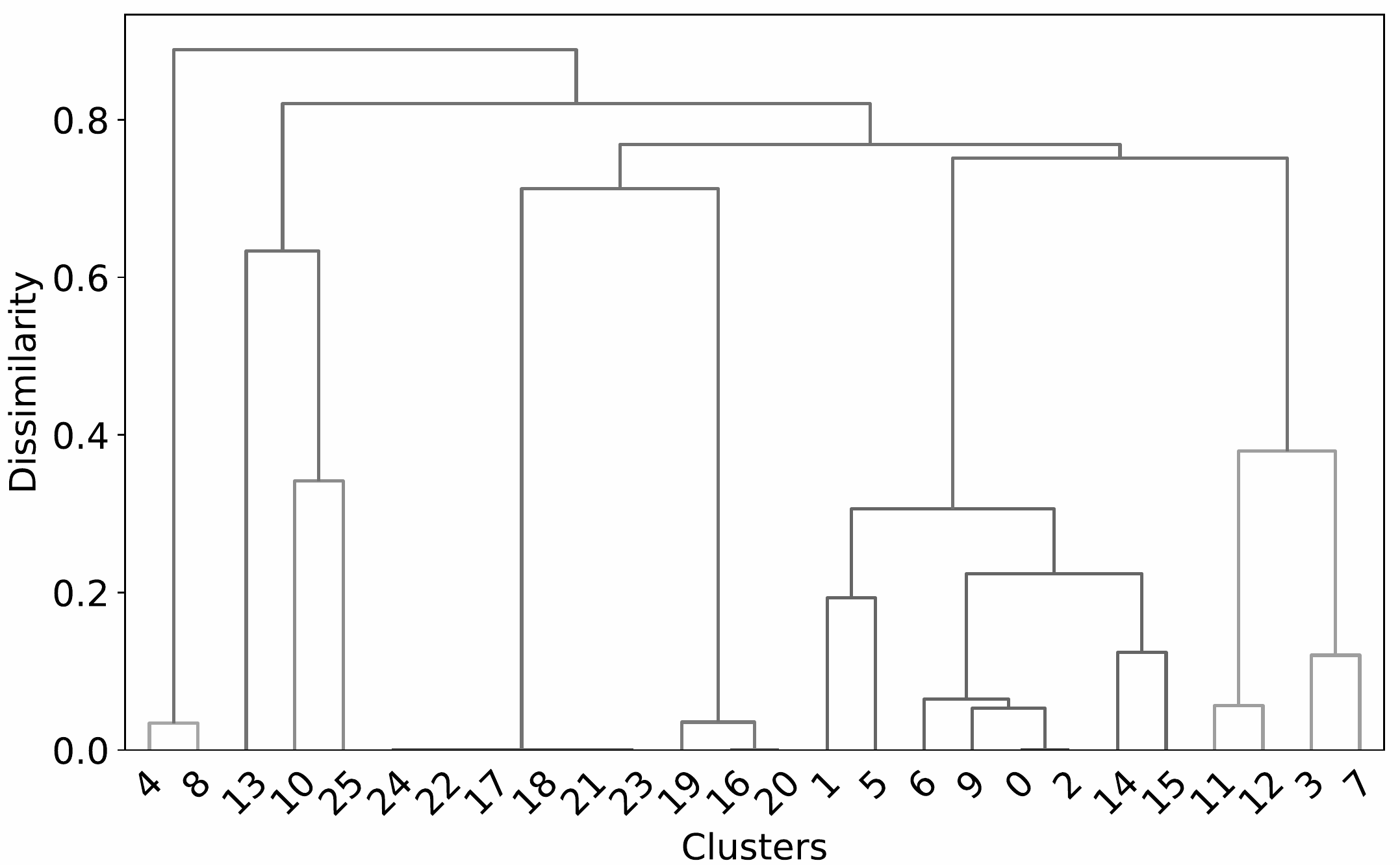}
    \caption{Agglomerative clustering of classification metrics  
    (using Spearman rank correlation). Here x-axis shows the classification metric Ids from Table~\ref{tbl:fairness_def} and y-axis shows the dissimilarity measure between clusters.}
    \label{fig:cm_dendo}
\end{figure}

Figure~\ref{fig:cm_dendo} shows the dendrogram created for the classification metrics using the above described method. Table~\ref{26_classification_metrics} shows that we get seven clusters from 26 classification metrics. Following a similar process for dataset metrics we get three clusters as shown in Table~\ref{tbl:dataset_metrics}.

To build these clusters and dendrograms, we measure the similarity of two metrics. In this paper,  by ``similarity'' we  mean, they  are measuring the similar bias in the models/dataset. Such similar metrics will show a similar pattern of changes in bias when models are built using different parts of the data or different bias removal algorithms are used. 
To compute this similarity, we sample from our model training procedure (see \S\ref{train}) that computes our metrics 25 times, each time using different train/validation/test samples of the data. Next, for each dataset, for those 25 numbers, we use correlation to assess similarity.

Two widely used definitions of correlation~\cite{dickinson2001finding, shirkhorshidi2015comparison, reeb2015assessing, do2008clustering, hossen2015methods, d2005does,  zhang2016cross, chen2017analytics} are the   (a)~Pearson correlation (which evaluates the linear relationship between two continuous variables) and the (b)~Spearman rank correlation (which is a non-parametric measure of rank correlation that evaluates the monotonic relationship between two continuous or ordinal variables).  We choose Spearman rank correlation, as it measures the monotonic relationship between two variables and is less affected by outliers. 

\subsection{Experimental Setup}
\label{subsec:framework}
We summarize our experimental setup as follows.

\subsubsection{\textbf{Data Pre-processing}:} Three different pre-processing steps are performed before using the data~\cite{lahoti2019ifair, friedler2019comparative, schelter2019fairprep} for model building. At first, each categorical value in the dataset is converted either using a label encoder or by one hot encoder, as most most ML algorithms can't handle categorical values directly. Then the protected attributes are changed into ones and zeros from their original values. Here we denote the privileged attribute as one and unprivileged as zero. Finally, we use min-max normalization in the datasets to normalize the data before building the models.

\subsubsection{\textbf{Model Training}:}\label{train} We used five fold cross-validation repeated five times with random seeds build training/ test sets (as recommended by \cite{verma2018fairness, schelter2019fairprep, valentim2019impact, kamiran2012data}). This step is to divide the data into multiple subsets of data with various degrees of bias. We train three models in each iteration (a)~\textit{baseline model:} here we use the training data to build a logistic regression model; (b)~\textit{Reweighing model:} here we first train the  reweighing method, then use the learned model to transform the training data to achieve group fairness. Using the transformed data, we train a logistic regression from scikit-learn with `l2' regularization, `lbfgs' solver and maximum iteration of 1000; and (c)~\textit{Meta Fair Classifier model:} here to train the meta fair classifier model, we use the training data to build multiple meta fair classifier model with different values of $\tau$ (a hyperparameter for fairness penalty in the model) and measure the bias in the model using the validation set. Then to build the final model, we select the $\tau$ for which the model had the lowest bias in the validation set and build the final meta fair classifier model. 
    
\subsubsection{\textbf{Fairness Metric Calculation}:} We collect the performance of each model based on 26 classification and four dataset metrics for each iteration of the cross-validation. So for each iteration, we use the test data for prediction and then use the predicted values along with the ground truth to calculate the 26 classification metrics. Similarly, we collect the four dataset metrics on the baseline and reweighing method. Meta fair classifier is not applicable in the case of dataset metrics. 
    
\subsubsection{\textbf{Measure for Fairness}:} Data Pre-processing, Model Training, and Fairness Metric Calculation steps are performed for each of seven datasets with five fold five repeat cross-validation. Then to measure if the model built on a dataset is fair or unfair according to a metric, we selected a threshold for each of the metrics. As mentioned in \S\ref{sec:Terminology}, that threshold is the \textit{fair range}. If a metric value falls in that range, we say it ``fair'' otherwise ``unfair''.
    
\subsubsection{\textbf{Building Clusters}:} One of the main goals of this study is to group a set of metrics together that perform similarly and measure similar kinds of bias. We use 26 classification metrics calculated on seven datasets with three different methods to calculate metric to metric correlation based on Spearman rank correlation coefficient. We do the same for the four dataset metrics as well. This provides us two correlation matrices: one 26x26  and one 4x4. After that, to build the clusters using the agglomerative clustering, we convert the similarity matrix into a dissimilarity matrix~\cite{hossen2015methods, d2005does} using equation~\ref{eq:dissimialrity}. We use this dissimilarity matrix to create the clusters. The agglomerative clustering process creates a dendrogram as shown in Figure~\ref{fig:cm_dendo}. Now to select the number of clusters, we cut the dendrogram at a height, where the clusters will remain unchanged with the most increase/decrease of the cutting threshold. For classification metrics, we cut the dendrogram (Figure~\ref{fig:cm_dendo}) at 0.57 as the clusters will remain unchanged between the cutoff value 0.49 and 0.64. Finally, we get the clusters containing classification metrics measuring similar kinds of bias. We perform the same process for dataset metrics and cut the dendrogram at a height of 0.4.
    
\begin{equation}
\label{eq:dissimialrity}
    d(x,y) = 1 - |sim(x,y)|
\end{equation}
    
\subsubsection{\textbf{Calculating Sensitivity}:} Research question four asks about the consistency of the metric values for three cases: (a) raw data, b) after applying Reweighing (RW), (c) after applying Meta Fair Classifier (MFC). As we are using five cross fold five repeats for all the datasets, we get 25 results for each dataset and report
for all seven datasets:
\begin{itemize}
\item the median  value: the 50th percentile (or $Q_2$);
\item the IQR: the (75-25)th percentile (or $Q_{3} - Q_{1}$) 
\end{itemize}

\section{Results}
\label{sec:results}
Our results are organized based on four research questions. 

\begin{RQ}
{\bf RQ1: Do current fairness metrics agree with each other?} 
\end{RQ}
At first, we need to verify our motivation. In real life, do the fairness metrics contradict?  Table~\ref{26_classification_metrics} contains results for 26 classification metrics;  Table~\ref{tbl:dataset_metrics} contains results for four dataset metrics. The learner here is logistic regression. The last row contains the percentage of metrics marking the specific dataset as unfair in both tables. If we combine last rows of Table~\ref{26_classification_metrics} \& ~\ref{tbl:dataset_metrics} and sort them in ascending order, we get the following list:


\begin{center}
           \{ 23, 34, 50, 50, 50, 54, 58, 65, 75, 75, 75, 75, 77, 100 \}\%
\end{center}

The median value here is 62\%; i.e., nearly half the time the metrics  make {\em different} conclusions about the {\em same} data. This means that researchers and practitioners will be spending much effort trying to understand their systems using disagreeing oracles (a result that  motivates this entire paper).

\begin{table}
\caption{Cluster based results for 26 classification metrics on seven datasets. For a metric with ideal an value of zero, anything below -0.1 and above 0.1 is ``unfair''. For a metric with an ideal value of one, anything <0.8 or >1.2 is ``unfair''.}
\label{26_classification_metrics}
\centering
\resizebox{\textwidth}{!}{%
\begin{tabular}{c|c|l|ccccccc|c}
\rowcolor[HTML]{C0C0C0} 
\multicolumn{1}{c|}{\cellcolor[HTML]{C0C0C0}}                                                                        & \multicolumn{1}{c|}{\cellcolor[HTML]{C0C0C0}}                      & \multicolumn{1}{c|}{\cellcolor[HTML]{C0C0C0}}                          & \multicolumn{7}{c|}{\cellcolor[HTML]{C0C0C0}Datasets}                                                                                                                                                                                                                                                                                                                                  & \cellcolor[HTML]{C0C0C0}                                                                                                      \\
\rowcolor[HTML]{C0C0C0} 
\multicolumn{1}{c|}{\cellcolor[HTML]{C0C0C0}}                                                                        & \multicolumn{1}{c|}{\cellcolor[HTML]{C0C0C0}}                      & \multicolumn{1}{c|}{\cellcolor[HTML]{C0C0C0}}                          & \cellcolor[HTML]{C0C0C0}                        & \cellcolor[HTML]{C0C0C0}                         & \cellcolor[HTML]{C0C0C0}                         & \cellcolor[HTML]{C0C0C0}                         & \cellcolor[HTML]{C0C0C0}                       & \cellcolor[HTML]{C0C0C0}                          & \multicolumn{1}{c|}{\cellcolor[HTML]{C0C0C0}}                          & \cellcolor[HTML]{C0C0C0}                                                                                                      \\
\rowcolor[HTML]{C0C0C0} 
\multicolumn{1}{c|}{\multirow{-3}{*}{\cellcolor[HTML]{C0C0C0}\begin{tabular}[c]{@{}c@{}}Cluster \\ Id\end{tabular}}} & \multicolumn{1}{c|}{\multirow{-3}{*}{\cellcolor[HTML]{C0C0C0}MID}} & \multicolumn{1}{c|}{\multirow{-3}{*}{\cellcolor[HTML]{C0C0C0}Metrics}} & \multirow{-2}{*}{\cellcolor[HTML]{C0C0C0}Adult} & \multirow{-2}{*}{\cellcolor[HTML]{C0C0C0}Compas} & \multirow{-2}{*}{\cellcolor[HTML]{C0C0C0}German} & \multirow{-2}{*}{\cellcolor[HTML]{C0C0C0}Health} & \multirow{-2}{*}{\cellcolor[HTML]{C0C0C0}Bank} & \multirow{-2}{*}{\cellcolor[HTML]{C0C0C0}Student} & \multicolumn{1}{c|}{\multirow{-2}{*}{\cellcolor[HTML]{C0C0C0}Titanic}} & \multirow{-3}{*}{\cellcolor[HTML]{C0C0C0}\begin{tabular}[c]{@{}c@{}}Metric\\ Type\end{tabular}}                               \\ \hline
\rowcolor[HTML]{FFFFFF} 
0                                                                                                                    & C3                                                                 & false\_omission\_rate\_difference                                      & Unfair                                          & Fair                                             & Fair                                             & Unfair                                           & Fair                                           & Fair                                              & Unfair                                                                 & \cellcolor[HTML]{FFFFFF}                                                                                                      \\
\rowcolor[HTML]{FFFFFF} 
0                                                                                                                    & C7                                                                 & false\_omission\_rate\_ratio                                           & Unfair                                          & Fair                                             & Fair                                             & Unfair                                           & Fair                                           & Unfair                                            & Unfair                                                                 & \cellcolor[HTML]{FFFFFF}                                                                                                      \\
\rowcolor[HTML]{FFFFFF} 
0                                                                                                                    & C11                                                                & error\_rate\_difference                                                & Unfair                                          & Fair                                             & Fair                                             & Unfair                                           & Fair                                           & Fair                                              & Fair                                                                   & \cellcolor[HTML]{FFFFFF}                                                                                                      \\
\rowcolor[HTML]{FFFFFF} 
0                                                                                                                    & C12                                                                & error\_rate\_ratio                                                     & Unfair                                          & Fair                                             & Fair                                             & Unfair                                           & Fair                                           & Fair                                              & Fair                                                                   &                        \\
\rowcolor[HTML]{FFFFFF} 
\multicolumn{3}{c}{\cellcolor[HTML]{FFFFFF}\textbf{Percentage of agreement}}                                                                                                                                                                        & \textbf{100\%}                                            & \textbf{100\%}                                             & \textbf{100\%}                                            & \textbf{100\%}                                             & \textbf{100\%}                                           & \textbf{75\%}                                              & \textbf{50\%} &  \multirow{-5}{*}{\cellcolor[HTML]{FFFFFF}\begin{tabular}[c]{@{}c@{}}Mis-\\ classification\end{tabular}}                       \\ \hline
\rowcolor[HTML]{FFCCC9} 
1                                                                                                                    & C10                                                                & average\_abs\_odds\_difference                                         & Unfair                                          & Unfair                                           & Unfair                                           & Unfair                                           & Unfair                                         & Fair                                              & Unfair                                                                 & \cellcolor[HTML]{FFCCC9}                                                                                                      \\
\rowcolor[HTML]{FFCCC9} 
1                                                                                                                    & C25                                                                & differential\_fairness\_bias\_amplification                            & Unfair                                          & Unfair                                           & Unfair                                           & Unfair                                           & Unfair                                         & Fair                                              & Unfair                                                                 & \\
\rowcolor[HTML]{FFCCC9} 
\multicolumn{3}{c}{\cellcolor[HTML]{FFCCC9}\textbf{Percentage of agreement}}                                                                                                                                                                        & \textbf{100\%}                                            & \textbf{100\%}                                             & \textbf{100\%}                                            & \textbf{100\%}                                             & \textbf{100\%}                                           & \textbf{100\%}                                              & \textbf{100\%} &  \multirow{-3}{*}{\cellcolor[HTML]{FFCCC9}\begin{tabular}[c]{@{}c@{}}Differential \\ Fairness\end{tabular}}                    \\ \hline
\rowcolor[HTML]{FFFFFF} 
2                                                                                                                    & C16                                                                & generalized\_entropy\_index                                            & Fair                                            & Unfair                                           & Fair                                             & Fair                                             & Fair                                           & Fair                                              & Unfair                                                                 & \cellcolor[HTML]{FFFFFF}                                                                                                      \\
\rowcolor[HTML]{FFFFFF} 
2                                                                                                                    & C19                                                                & theil\_index                                                           & Unfair                                          & Unfair                                           & Fair                                             & Unfair                                           & Unfair                                         & Fair                                              & Unfair                                                                 & \cellcolor[HTML]{FFFFFF}                                                                                                      \\
\rowcolor[HTML]{FFFFFF} 
2                                                                                                                    & C20                                                                & coefficient\_of\_variation                                             & Unfair                                          & Unfair                                           & Unfair                                           & Unfair                                           & Unfair                                         & Unfair                                            & Unfair                                                                 & \\
\rowcolor[HTML]{FFFFFF} 
\multicolumn{3}{c}{\cellcolor[HTML]{FFFFFF}\textbf{Percentage of agreement}}                                                                                                                                                                        & \textbf{67\%}                                            & \textbf{100\%}                                             & \textbf{67\%}                                            & \textbf{67\%}                                             & \textbf{67\%}                                           & \textbf{67\%}                   &   \textbf{100\%}                         & \multirow{-4}{*}{\cellcolor[HTML]{FFFFFF}\begin{tabular}[c]{@{}c@{}}Individual \\ Fairness\end{tabular}}                      \\ \hline
\rowcolor[HTML]{FFCCC9} 
3                                                                                                                    & C4                                                                 & false\_discovery\_rate\_difference                                     & Fair                                            & Fair                                             & Fair                                             & Fair                                             & Fair                                           & Fair                                              & Unfair                                                                 & \cellcolor[HTML]{FFCCC9}                                                                                                      \\
\rowcolor[HTML]{FFCCC9} 
3                                                                                                                    & C8                                                                 & false\_discovery\_rate\_ratio                                          & Fair                                            & Fair                                             & Fair                                             & Fair                                             & Fair                                           & Unfair                                            & Unfair                                                                 & \\
\rowcolor[HTML]{FFCCC9} 
\multicolumn{3}{c}{\cellcolor[HTML]{FFCCC9}\textbf{Percentage of agreement}}                                                                                                                                                                        & \textbf{100\%}                                            & \textbf{100\%}                                             & \textbf{100\%}                                            & \textbf{65\%}                                             & \textbf{100\%}                                           & \textbf{50\%}                   &   \textbf{100\%}                         &  \multirow{-3}{*}{\cellcolor[HTML]{FFCCC9}\begin{tabular}[c]{@{}c@{}}Mis-\\ classification\end{tabular}}                       \\ \hline
\rowcolor[HTML]{FFFFFF} 
4                                                                                                                    & C0                                                                 & true\_positive\_rate\_difference                                       & Unfair                                          & Unfair                                           & Fair                                             & Unfair                                           & Unfair                                         & Fair                                              & Unfair                                                                 & \cellcolor[HTML]{FFFFFF}                                                                                                      \\
\rowcolor[HTML]{FFFFFF} 
4                                                                                                                    & C1                                                                 & false\_positive\_rate\_difference                                      & Fair                                            & Unfair                                           & Unfair                                           & Unfair                                           & Unfair                                         & Fair                                              & Unfair                                                                 & \cellcolor[HTML]{FFFFFF}                                                                                                      \\
\rowcolor[HTML]{FFFFFF} 
4                                                                                                                    & C2                                                                 & false\_negative\_rate\_difference                                      & Unfair                                          & Unfair                                           & Unfair                                           & Unfair                                           & Unfair                                         & Fair                                              & Unfair                                                                 & \cellcolor[HTML]{FFFFFF}                                                                                                      \\
\rowcolor[HTML]{FFFFFF} 
4                                                                                                                    & C5                                                                 & false\_positive\_rate\_ratio                                           & Fair                                            & Unfair                                           & Unfair                                           & Unfair                                           & Unfair                                         & Fair                                              & Unfair                                                                 & \cellcolor[HTML]{FFFFFF}                                                                                                      \\
\rowcolor[HTML]{FFFFFF} 
4                                                                                                                    & C6                                                                 & false\_negative\_rate\_ratio                                           & Unfair                                          & Unfair                                           & Unfair                                           & Unfair                                           & Unfair                                         & Unfair                                            & Unfair                                                                 & \cellcolor[HTML]{FFFFFF}                                                                                                      \\
\rowcolor[HTML]{FFFFFF} 
4                                                                                                                    & C9                                                                 & average\_odds\_difference                                              & Unfair                                          & Unfair                                           & Unfair                                           & Unfair                                           & Unfair                                         & Fair                                              & Unfair                                                                 & \cellcolor[HTML]{FFFFFF}                                                                                                      \\
\rowcolor[HTML]{FFFFFF} 
4                                                                                                                    & C14                                                                & disparate\_impact                                                      & Unfair                                          & Unfair                                           & Unfair                                           & Unfair                                           & Unfair                                         & Unfair                                            & Unfair                                                                 & \cellcolor[HTML]{FFFFFF}                                                                                                      \\
\rowcolor[HTML]{FFFFFF} 
4                                                                                                                    & C15                                                                & statistical\_parity\_difference                                        & Unfair                                          & Unfair                                           & Unfair                                           & Unfair                                           & Unfair                                         & Fair                                              & Unfair                                                                 & \\
\rowcolor[HTML]{FFFFFF} 
\multicolumn{3}{c}{\cellcolor[HTML]{FFFFFF}\textbf{Percentage of agreement}}                                                                                                                                                                        & \textbf{75\%}                                            & \textbf{100\%}                                             & \textbf{88\%}                                            & \textbf{100\%}                                             & \textbf{100\%}                                           & \textbf{75\%}                   &   \textbf{100\%}                         & \multirow{-9}{*}{\cellcolor[HTML]{FFFFFF}\begin{tabular}[c]{@{}c@{}}Confusion \\ Matrix Based \\ Group Fairness\end{tabular}} \\ \hline
\rowcolor[HTML]{FFCCC9} 
5                                                                                                                    & C17                                                                & between\_all\_groups\_generalized\_entropy\_index                      & Fair                                            & Fair                                             & Fair                                             & Fair                                             & Fair                                           & Fair                                              & Fair                                                                   & \cellcolor[HTML]{FFCCC9}                                                                                                      \\
\rowcolor[HTML]{FFCCC9} 
5                                                                                                                    & C18                                                                & between\_group\_generalized\_entropy\_index                            & Fair                                            & Fair                                             & Fair                                             & Fair                                             & Fair                                           & Fair                                              & Fair                                                                   & \cellcolor[HTML]{FFCCC9}                                                                                                      \\
\rowcolor[HTML]{FFCCC9} 
5                                                                                                                    & C21                                                                & between\_group\_theil\_index                                           & Fair                                            & Fair                                             & Fair                                             & Fair                                             & Fair                                           & Fair                                              & Fair                                                                   & \cellcolor[HTML]{FFCCC9}                                                                                                      \\
\rowcolor[HTML]{FFCCC9} 
5                                                                                                                    & C22                                                                & between\_group\_coefficient\_of\_variation                             & Fair                                            & Fair                                             & Fair                                             & Fair                                             & Fair                                           & Fair                                              & Unfair                                                                 & \cellcolor[HTML]{FFCCC9}                                                                                                      \\
\rowcolor[HTML]{FFCCC9} 
5                                                                                                                    & C23                                                                & between\_all\_groups\_theil\_index                                     & Fair                                            & Fair                                             & Fair                                             & Fair                                             & Fair                                           & Fair                                              & Fair                                                                   & \cellcolor[HTML]{FFCCC9}                                                                                                      \\
\rowcolor[HTML]{FFCCC9} 
5                                                                                                                    & C24                                                                & between\_all\_groups\_coefficient\_of\_variation                       & Fair                                            & Fair                                             & Fair                                             & Fair                                             & Fair                                           & Fair                                              & Unfair                                                                 & \\
\rowcolor[HTML]{FFCCC9} 
\multicolumn{3}{c}{\cellcolor[HTML]{FFCCC9}\textbf{Percentage of agreement}}                                                                                                                                                                        & \textbf{100\%}                                            & \textbf{100\%}                                             & \textbf{100\%}                                            & \textbf{100\%}                                             & \textbf{100\%}                                           & \textbf{100\%}                   &   \textbf{67\%}                         & \multirow{-7}{*}{\cellcolor[HTML]{FFCCC9}\begin{tabular}[c]{@{}c@{}}Between \\ Group \\ Individual \\ Fairness\end{tabular}}  \\ \hline
\rowcolor[HTML]{FFFFFF} 
6                                                                                                                    & C13                                                                & selection\_rate                                                        & Unfair                                          & Unfair                                           & Unfair                                           & Unfair                                           & Unfair                                         & Unfair                                            & Unfair                                                                 & \\
\rowcolor[HTML]{FFFFFF} 
\multicolumn{3}{c}{\cellcolor[HTML]{FFFFFF}\textbf{Percentage of agreement}}                                                                                                                                                                        & \textbf{100\%}                                            & \textbf{100\%}                                             & \textbf{100\%}                                            & \textbf{100\%}                                             & \textbf{100\%}                                           & \textbf{100\%}                   &   \textbf{100\%}                         & \begin{tabular}[c]{@{}c@{}}Intermediate \\ Metric\end{tabular}                                                                \\ \hline
\rowcolor[HTML]{C0C0C0} 
\multicolumn{3}{c}{\cellcolor[HTML]{C0C0C0}\textbf{Percentage of metrics marking dataset as unfair}}                                                                                                                                                                        & \textbf{58\%}                                            & \textbf{54\%}                                             & \textbf{34\%}                                            & \textbf{65\%}                                             & \textbf{50\%}                                           & \textbf{23\%}                                              & \textbf{77\%}                                                                   &                                                                                                                              
\end{tabular}}
\bigskip
\caption{Cluster based results for four dataset metrics on seven datasets. For a metric with ideal value of zero, anything below -0.1 and above 0.1 is ``unfair''. For a metric with ideal value of one, anything <0.8 or >1.2 is ``unfair''.}
\label{tbl:dataset_metrics}
\resizebox{\textwidth}{!}{%
\begin{tabular}{c|c|l|ccccccc|c}
\rowcolor[HTML]{C0C0C0} 
\cellcolor[HTML]{C0C0C0}                             & \cellcolor[HTML]{C0C0C0}                      & \multicolumn{1}{c|}{\cellcolor[HTML]{C0C0C0}}                          & \multicolumn{7}{c|}{\cellcolor[HTML]{C0C0C0}Datasets}                                                                                                                                                                                                                                                                                                             & \cellcolor[HTML]{C0C0C0}                                                                                                      \\
\rowcolor[HTML]{C0C0C0} 
\cellcolor[HTML]{C0C0C0}                             & \cellcolor[HTML]{C0C0C0}                      & \multicolumn{1}{c|}{\cellcolor[HTML]{C0C0C0}}                          & \cellcolor[HTML]{C0C0C0}                        & \cellcolor[HTML]{C0C0C0}                         & \cellcolor[HTML]{C0C0C0}                         & \cellcolor[HTML]{C0C0C0}                         & \cellcolor[HTML]{C0C0C0}                       & \cellcolor[HTML]{C0C0C0}                          & \cellcolor[HTML]{C0C0C0}                          & \cellcolor[HTML]{C0C0C0}                                                                                                      \\
\rowcolor[HTML]{C0C0C0} 
\multirow{-3}{*}{\cellcolor[HTML]{C0C0C0}Cluster Id} & \multirow{-3}{*}{\cellcolor[HTML]{C0C0C0}MID} & \multicolumn{1}{c|}{\multirow{-3}{*}{\cellcolor[HTML]{C0C0C0}Metrics}} & \multirow{-2}{*}{\cellcolor[HTML]{C0C0C0}Adult} & \multirow{-2}{*}{\cellcolor[HTML]{C0C0C0}Compas} & \multirow{-2}{*}{\cellcolor[HTML]{C0C0C0}German} & \multirow{-2}{*}{\cellcolor[HTML]{C0C0C0}Health} & \multirow{-2}{*}{\cellcolor[HTML]{C0C0C0}Bank} & \multirow{-2}{*}{\cellcolor[HTML]{C0C0C0}Student} & \multirow{-2}{*}{\cellcolor[HTML]{C0C0C0}Titanic} & \multirow{-3}{*}{\cellcolor[HTML]{C0C0C0}\begin{tabular}[c]{@{}c@{}}Metric\\ Type\end{tabular}}                               \\ \hline
\rowcolor[HTML]{FFCCC9} 
0                                                    & D0                                            & consistency                                                            & Fair                                            & Unfair                                           & Fair                                             & Unfair                                           & Fair                                           & Unfair                                            & Fair                                              & Individual Fairness                                                                                                           \\ \hline
\rowcolor[HTML]{FFFFFF} 
1                                                    & D1                                            & smoothed\_empirical\_differential\_fairness                            & Unfair                                          & Unfair                                           & Unfair                                           & Unfair                                           & Unfair                                         & Unfair                                            & Unfair                                            & Differential Fairness                                                                                                         \\ \hline
\rowcolor[HTML]{FFCCC9} 
2                                                    & D2                                            & mean\_difference                                                       & Unfair                                          & Unfair                                           & Unfair                                           & Fair                                             & Unfair                                         & Fair                                              & Unfair                                            & \cellcolor[HTML]{FFCCC9}                                                                                                      \\
\rowcolor[HTML]{FFCCC9} 
2                                                    & D3                                            & disparate\_impact                                                      & Unfair                                          & Unfair                                           & Unfair                                           & Fair                                             & Unfair                                         & Fair                                              & Unfair                                            & \cellcolor[HTML]{FFCCC9}                                                                                                      \\
\rowcolor[HTML]{FFCCC9} 
\multicolumn{1}{l|}{\cellcolor[HTML]{FFCCC9}}        & \multicolumn{1}{l|}{\cellcolor[HTML]{FFCCC9}} &                                                                        & \multicolumn{1}{l}{\cellcolor[HTML]{FFCCC9}}    & \multicolumn{1}{l}{\cellcolor[HTML]{FFCCC9}}     & \multicolumn{1}{l}{\cellcolor[HTML]{FFCCC9}}     & \multicolumn{1}{l}{\cellcolor[HTML]{FFCCC9}}     & \multicolumn{1}{l}{\cellcolor[HTML]{FFCCC9}}   & \multicolumn{1}{l}{\cellcolor[HTML]{FFCCC9}}      & \multicolumn{1}{l|}{\cellcolor[HTML]{FFCCC9}}     & \multirow{-3}{*}{\cellcolor[HTML]{FFCCC9}\begin{tabular}[c]{@{}c@{}}Confusion \\ Matrix Based \\ Group Fairness\end{tabular}} \\ \hline
\rowcolor[HTML]{C0C0C0} 
\multicolumn{3}{c}{\cellcolor[HTML]{C0C0C0}\textbf{Percentage of metrics marking dataset as unfair}}                                                                                                                                                                        & \textbf{75\%}                                            & \textbf{100\%}                                             & \textbf{75\%}                                             & \textbf{50\%}                                             & \textbf{75\%}                                           & \textbf{50\%}                                              & \textbf{75\%}    & 
\end{tabular}}
\end{table}

\begin{RQ}
{\bf RQ2: Can we group (cluster) fairness metrics based on similarity?} 
\end{RQ}


Table~\ref{26_classification_metrics} shows that 26 classification metrics can be divided into seven clusters. Table~\ref{tbl:dataset_metrics} shows that four dataset metrics can be divided into three clusters.  More importantly, we note that:


\begin{itemize}
\item RQ1 reported intra-project disagreement on ``fair``-vs-``unfair'';
\item We note that there is  much intra-cluster agreement for each data set in  Table~\ref{26_classification_metrics} and Table~\ref{tbl:dataset_metrics}.
\end{itemize}

As evidence, we note that the majority fairness decision is always the same within the clusters for each dataset. In Table~\ref{26_classification_metrics}, the row {\em Percentage of agreement} comments on the uniformity of decisions within each cluster (for each dataset). Note that uniformity is very high (often 100\%). That means metrics inside each cluster agree with each other for every dataset. Among the seven clusters, we see six clusters (except cluster two) show 100\% agreement considering median value across seven datasets. For example, in case of cluster zero, percentage of agreement is 100\% for five datasets; 75\% for one; 50\% for one. Majority is 100\%. That is true for clusters 1,3,4,5,6 \& 7. We see similar agreement pattern inside clusters in Table~\ref{tbl:dataset_metrics} also.


For reference purposes, the last column of Table~\ref{26_classification_metrics} and Table~\ref{tbl:dataset_metrics} offers names for those clusters:
\begin{itemize}
    \item  \textbf{ Misclassification (cluster 0, 3)}: these metrics try to measure the difference or ratio of misclassification errors between groups;
    \item \textbf{Differential fairness (cluster 1)}: these metrics try to measure if probabilities of the outcomes are similar regardless of the combination of protected attributes~\cite{Foulds2019DifferentialF}; 
    \item \textbf{Individual Fairness  (cluster 2)}: It measures if two similar individuals with respect to the classification task receive the same outcome or not; 
    \item \textbf{ Confusion matrix based group fairness (cluster 4)}: these metrics measure difference or ratio between groups based on confusion matrix; 
    \item \textbf{Between group individual fairness (cluster 5)}: measures the difference or ratio of individual fairness between groups; 
    \item \textbf{Intermediate metrics (cluster 6)}: these are intermediate metrics. 
\end{itemize}
From a practitioner viewpoint, this clustering is useful because:
\begin{itemize}
    \item The clustering reduces the confusion of having too many metrics and not knowing their similarity.
    \item As the metrics inside the same cluster measure same kind of bias and behave in the same manner; we can choose just one metric from each cluster. Thus we measure a few metrics but can cover a much more comprehensive range of fairness notions.
    \item If we see agreement among all the metrics inside a cluster for a particular dataset, then one metric can be chosen as representative of the whole cluster.
    \item In case of intra-cluster conflicts, choosing only one metric can be risky. In these cases, practitioners need to do a proper risk assessment before selecting metrics. That said, if there is intra-cluster conflict among metrics, we can choose one from the `fair' group and one from the `unfair' group to mitigate that risk.
\end{itemize} 

As part of this study, we further analyzed each cluster mathematically to verify if our cluster of metrics and their mathematical definitions coincide. A detailed analysis of these clusters and their mathematical analysis has been discussed in \S\ref{sec:cluster_analysis}.

\begin{table}[!b]
\caption{This table shows sensitivity of the classification metrics on the three different models used in this study (a) Baseline; (b) Reweighing(RW); and (c) Meta Fair Classifier(MFC). The table shows the median and IQR values of three datasets.  Here the cells in IQR columns are marked with ``red'' those that change by more than a small amount (35th percentile of the standard deviation of the IQR values). The insensitive metrics are those that usually have white IQR values. }
\label{tbl:rq4}
\resizebox{\textwidth}{!}{%
\begin{tabular}{c|cccccc|cccccc|cccccc}
\rowcolor[HTML]{C0C0C0} 
\cellcolor[HTML]{C0C0C0}                      & \multicolumn{6}{c}{\cellcolor[HTML]{C0C0C0}Compas}                                                                                                                & \multicolumn{6}{c|}{\cellcolor[HTML]{C0C0C0}Health}                                                                                                                & \multicolumn{6}{c|}{\cellcolor[HTML]{C0C0C0}German}                                                                                                      \\
\rowcolor[HTML]{C0C0C0} 
\cellcolor[HTML]{C0C0C0}                      & \multicolumn{2}{c}{\cellcolor[HTML]{C0C0C0}Baseline} & \multicolumn{2}{c}{\cellcolor[HTML]{C0C0C0}RW} & \multicolumn{2}{c}{\cellcolor[HTML]{C0C0C0}MFC}           & \multicolumn{2}{c|}{\cellcolor[HTML]{C0C0C0}Baseline} & \multicolumn{2}{c}{\cellcolor[HTML]{C0C0C0}RW} & \multicolumn{2}{c}{\cellcolor[HTML]{C0C0C0}MFC}           & \multicolumn{2}{c|}{\cellcolor[HTML]{C0C0C0}Baseline} & \multicolumn{2}{c}{\cellcolor[HTML]{C0C0C0}RW} & \multicolumn{2}{c}{\cellcolor[HTML]{C0C0C0}MFC} \\
\rowcolor[HTML]{C0C0C0} 
\multirow{-3}{*}{\cellcolor[HTML]{C0C0C0}MID} & Med           & IQR                                  & Med        & IQR                               & Med    & \multicolumn{1}{c|}{\cellcolor[HTML]{C0C0C0}IQR} & Med            & IQR                                  & Med        & IQR                               & Med    & \multicolumn{1}{c|}{\cellcolor[HTML]{C0C0C0}IQR} & Med            & IQR                                  & Med        & IQR                               & Med         & IQR                               \\ \hline
C3                                            & -0.067        & \cellcolor[HTML]{FFCCC9}0.079        & -0.113     & 0.015                             & -0.061 & \cellcolor[HTML]{FFCCC9}0.035                    & -0.14          & 0.032                                & -0.211     & \cellcolor[HTML]{FFCCC9}0.068     & -0.151 & \cellcolor[HTML]{FFCCC9}0.137                    & 0              & \cellcolor[HTML]{FFCCC9}0.5          & -0.5       & \cellcolor[HTML]{FFCCC9}0.667     & 0           & \cellcolor[HTML]{FFCCC9}0.592     \\
C7                                            & 0.817         & \cellcolor[HTML]{FFCCC9}0.211        & 0.691      & \cellcolor[HTML]{FFCCC9}0.029     & 0.824  & \cellcolor[HTML]{FFCCC9}0.099                    & 0.357          & \cellcolor[HTML]{FFCCC9}0.091        & 0.158      & \cellcolor[HTML]{FFCCC9}0.333     & 0.408  & \cellcolor[HTML]{FFCCC9}0.356                    & 2.3            & \cellcolor[HTML]{FFCCC9}0.72         & 0          & \cellcolor[HTML]{FFCCC9}0.5       & 1           & \cellcolor[HTML]{FFCCC9}0.53      \\
C11                                           & -0.033        & \cellcolor[HTML]{FFCCC9}0.043        & -0.016     & \cellcolor[HTML]{FFCCC9}0.039     & -0.042 & \cellcolor[HTML]{FFCCC9}0.026                    & -0.108         & 0.01                                 & -0.133     & 0.013                             & -0.098 & \cellcolor[HTML]{FFCCC9}0.111                    & 0.059          & \cellcolor[HTML]{FFCCC9}0.074        & 0.059      & \cellcolor[HTML]{FFCCC9}0.114     & 0.049       & \cellcolor[HTML]{FFCCC9}0.062     \\
C12                                           & 0.912         & \cellcolor[HTML]{FFCCC9}0.112        & 0.958      & \cellcolor[HTML]{FFCCC9}0.112     & 0.887  & \cellcolor[HTML]{FFCCC9}0.071                    & 0.508          & \cellcolor[HTML]{FFCCC9}0.174        & 0.339      & 0.026                             & 0.494  & \cellcolor[HTML]{FFCCC9}0.51                     & 1.18           & \cellcolor[HTML]{FFCCC9}0.274        & 1.18       & \cellcolor[HTML]{FFCCC9}0.418     & 1.166       & \cellcolor[HTML]{FFCCC9}0.215     \\ \hline
C10                                           & 0.252         & \cellcolor[HTML]{FFCCC9}0.058        & 0.029      & \cellcolor[HTML]{FFCCC9}0.02      & 0.181  & \cellcolor[HTML]{FFCCC9}0.035                    & 0.162          & \cellcolor[HTML]{FFCCC9}0.103        & 0.106      & \cellcolor[HTML]{FFCCC9}0.087     & 0.161  & \cellcolor[HTML]{FFCCC9}0.064                    & 0.221          & \cellcolor[HTML]{FFCCC9}0.167        & 0.043      & \cellcolor[HTML]{FFCCC9}0.048     & 0.031       & \cellcolor[HTML]{FFCCC9}0.121     \\
C25                                           & 0.531         & \cellcolor[HTML]{FFCCC9}0.354        & -0.22      & \cellcolor[HTML]{FFCCC9}0.141     & 0.359  & \cellcolor[HTML]{FFCCC9}0.148                    & 0.193          & \cellcolor[HTML]{FFCCC9}0.249        & -0.094     & \cellcolor[HTML]{FFCCC9}0.422     & 0.113  & \cellcolor[HTML]{FFCCC9}0.428                    & 2.399          & \cellcolor[HTML]{FFCCC9}3.29         & 1.162      & \cellcolor[HTML]{FFCCC9}0.49      & 1.578       & \cellcolor[HTML]{FFCCC9}2.087     \\ \hline
C16                                           & 0.193         & 0.001                                & 0.189      & 0.012                             & 0.192  & 0.007                                            & 0.091          & 0.004                                & 0.087      & 0.017                             & 0.091  & 0.025                                            & 0.076          & 0.011                                & 0.071      & 0.011                             & 0.066       & 0.011                             \\
C19                                           & 0.268         & 0.003                                & 0.263      & 0.017                             & 0.269  & 0.009                                            & 0.132          & 0.021                                & 0.14       & \cellcolor[HTML]{FFCCC9}0.051     & 0.139  & 0.033                                            & 0.083          & 0.02                                 & 0.073      & 0.017                             & 0.064       & 0.02                              \\
C20                                           & 0.878         & 0.003                                & 0.87       & \cellcolor[HTML]{FFCCC9}0.027     & 0.876  & 0.016                                            & 0.602          & 0.015                                & 0.589      & \cellcolor[HTML]{FFCCC9}0.058     & 0.602  & \cellcolor[HTML]{FFCCC9}0.082                    & 0.553          & 0.041                                & 0.532      & \cellcolor[HTML]{FFCCC9}0.041     & 0.513       & \cellcolor[HTML]{FFCCC9}0.043     \\ \hline
C4                                            & 0.04          & 0.034                                & 0.138      & \cellcolor[HTML]{FFCCC9}0.051     & 0.037  & \cellcolor[HTML]{FFCCC9}0.058                    & -0.091         & \cellcolor[HTML]{FFCCC9}0.133        & -0.009     & \cellcolor[HTML]{FFCCC9}0.202     & -0.016 & \cellcolor[HTML]{FFCCC9}0.152                    & 0.059          & \cellcolor[HTML]{FFCCC9}0.135        & 0.059      & \cellcolor[HTML]{FFCCC9}0.114     & 0.045       & \cellcolor[HTML]{FFCCC9}0.064     \\
C8                                            & 1.109         & \cellcolor[HTML]{FFCCC9}0.089        & 1.376      & \cellcolor[HTML]{FFCCC9}0.143     & 1.098  & \cellcolor[HTML]{FFCCC9}0.156                    & 0              & \cellcolor[HTML]{FFCCC9}0.944        & 0.964      & \cellcolor[HTML]{FFCCC9}1.571     & 0.925  & \cellcolor[HTML]{FFCCC9}1.292                    & 2.6            & \cellcolor[HTML]{FFCCC9}0.542        & 1.18       & \cellcolor[HTML]{FFCCC9}0.459     & 1.156       & \cellcolor[HTML]{FFCCC9}0.233     \\ \hline
C0                                            & -0.273        & \cellcolor[HTML]{FFCCC9}0.087        & -0.004     & \cellcolor[HTML]{FFCCC9}0.052     & -0.212 & \cellcolor[HTML]{FFCCC9}0.048                    & -0.106         & \cellcolor[HTML]{FFCCC9}0.139        & 0.13       & \cellcolor[HTML]{FFCCC9}0.227     & -0.117 & \cellcolor[HTML]{FFCCC9}0.405                    & -0.077         & \cellcolor[HTML]{FFCCC9}0.092        & 0          & \cellcolor[HTML]{FFCCC9}0.038     & -0.017      & \cellcolor[HTML]{FFCCC9}0.062     \\
C1                                            & -0.186        & \cellcolor[HTML]{FFCCC9}0.052        & -0.014     & \cellcolor[HTML]{FFCCC9}0.02      & -0.17  & \cellcolor[HTML]{FFCCC9}0.031                    & -0.214         & \cellcolor[HTML]{FFCCC9}0.053        & -0.13      & \cellcolor[HTML]{FFCCC9}0.174     & -0.109 & \cellcolor[HTML]{FFCCC9}0.114                    & -0.3           & \cellcolor[HTML]{FFCCC9}0.233        & 0          & 0.029                             & -0.053      & \cellcolor[HTML]{FFCCC9}0.176     \\
C2                                            & 0.273         & \cellcolor[HTML]{FFCCC9}0.087        & 0.004      & \cellcolor[HTML]{FFCCC9}0.052     & 0.212  & \cellcolor[HTML]{FFCCC9}0.048                    & 0.106          & \cellcolor[HTML]{FFCCC9}0.139        & -0.13      & \cellcolor[HTML]{FFCCC9}0.227     & 0.117  & \cellcolor[HTML]{FFCCC9}0.405                    & 0.077          & \cellcolor[HTML]{FFCCC9}0.092        & 0          & \cellcolor[HTML]{FFCCC9}0.038     & 0.017       & \cellcolor[HTML]{FFCCC9}0.062     \\
C5                                            & 0.408         & 0.037                                & 0.956      & \cellcolor[HTML]{FFCCC9}0.069     & 0.471  & \cellcolor[HTML]{FFCCC9}0.068                    & 0              & \cellcolor[HTML]{FFCCC9}0.219        & 0.25       & \cellcolor[HTML]{FFCCC9}0.654     & 0.191  & \cellcolor[HTML]{FFCCC9}0.347                    & 0.7            & \cellcolor[HTML]{FFCCC9}0.228        & 1          & 0.029                             & 0.947       & \cellcolor[HTML]{FFCCC9}0.176     \\
C6                                            & 1.71          & \cellcolor[HTML]{FFCCC9}0.242        & 1.009      & \cellcolor[HTML]{FFCCC9}0.124     & 1.514  & \cellcolor[HTML]{FFCCC9}0.133                    & 1.467          & \cellcolor[HTML]{FFCCC9}0.5          & 0.429      & \cellcolor[HTML]{FFCCC9}1.222     & 1.453  & \cellcolor[HTML]{FFCCC9}2.056                    & 3.4            & \cellcolor[HTML]{FFCCC9}0.56         & 0          & \cellcolor[HTML]{FFCCC9}5.459     & 11.532      & \cellcolor[HTML]{FFCCC9}3.4       \\
C9                                            & -0.252        & \cellcolor[HTML]{FFCCC9}0.058        & -0.018     & \cellcolor[HTML]{FFCCC9}0.059     & -0.181 & \cellcolor[HTML]{FFCCC9}0.035                    & -0.162         & \cellcolor[HTML]{FFCCC9}0.103        & -0.06      & \cellcolor[HTML]{FFCCC9}0.158     & -0.139 & \cellcolor[HTML]{FFCCC9}0.165                    & -0.221         & \cellcolor[HTML]{FFCCC9}0.167        & 0          & \cellcolor[HTML]{FFCCC9}0.043     & -0.031      & \cellcolor[HTML]{FFCCC9}0.121     \\
C14                                           & 0.432         & \cellcolor[HTML]{FFCCC9}0.073        & 0.89       & \cellcolor[HTML]{FFCCC9}0.133     & 0.541  & \cellcolor[HTML]{FFCCC9}0.062                    & 0.314          & \cellcolor[HTML]{FFCCC9}0.14         & 0.435      & \cellcolor[HTML]{FFCCC9}0.322     & 0.387  & \cellcolor[HTML]{FFCCC9}0.274                    & 0.836          & \cellcolor[HTML]{FFCCC9}0.12         & 1          & \cellcolor[HTML]{FFCCC9}0.045     & 0.971       & \cellcolor[HTML]{FFCCC9}0.105     \\
C15                                           & -0.264        & \cellcolor[HTML]{FFCCC9}0.05         & -0.049     & \cellcolor[HTML]{FFCCC9}0.059     & -0.205 & \cellcolor[HTML]{FFCCC9}0.03                     & -0.356         & \cellcolor[HTML]{FFCCC9}0.079        & -0.289     & \cellcolor[HTML]{FFCCC9}0.179     & -0.298 & \cellcolor[HTML]{FFCCC9}0.18                     & -0.164         & \cellcolor[HTML]{FFCCC9}0.122        & 0          & \cellcolor[HTML]{FFCCC9}0.043     & -0.029      & \cellcolor[HTML]{FFCCC9}0.104     \\ \hline
C17                                           & 0.002         & 0.002                                & 0.001      & 0.002                                 & 0.001  & 0.001                                            & 0.001              & 0.003                                    & 0.002          & 0.001                             & 0.001  & 0.001                                            & 0.001              & 0.002                                & 0.002          & 0.001                             & 0.001           & 0.001                             \\
C18                                           & 0.002         & 0.002                                & 0.001      & 0.002                                 & 0.001  & 0.001                                            & 0.001              & 0.001                                    & 0.002          & 0.001                             & 0.001  & 0.001                                            & 0.001              & 0.002                                & 0.002          & 0.001                             & 0.001           & 0.001                             \\
C21                                           & 0.002         & 0.002                                & 0.001      & 0.003                                 & 0.001  & 0.001                                            & 0.003              & 0.001                                    & 0.002          & 0.001                             & 0.001  & 0.001                                            & 0.003              & 0.002                                & 0.003          & 0.001                             & 0.002           & 0.001                             \\
C22                                           & 0.088         & \cellcolor[HTML]{FFCCC9}0.049        & 0.052      & 0.006                             & 0.057  & 0.019                                            & 0.036          & 0.037                                & 0.017      & \cellcolor[HTML]{FFCCC9}0.054     & 0.045  & 0.03                                             & 0.029          & \cellcolor[HTML]{FFCCC9}0.063        & 0.03       & \cellcolor[HTML]{FFCCC9}0.05      & 0.036       & 0.038                             \\
C23                                           & 0.002         & 0.002                                & 0.001      & 0                                 & 0.001  & 0.001                                            & 0.003              & 0.003                                    & 0.006          & 0.001                             & 0.001  & 0.001                                            & 0.001              & 0.002                                & 0.001          & 0.001                             & 0.004           & 0.001                             \\
C24                                           & 0.088         & \cellcolor[HTML]{FFCCC9}0.049        & 0.052      & 0.006                             & 0.057  & 0.019                                            & 0.036          & 0.037                                & 0.017      & \cellcolor[HTML]{FFCCC9}0.054     & 0.045  & 0.03                                             & 0.029          & \cellcolor[HTML]{FFCCC9}0.063        & 0.03       & \cellcolor[HTML]{FFCCC9}0.05      & 0.036       & 0.038                             \\ \hline
C13                                           & 0.405         & 0.025                                & 0.436      & 0.016                             & 0.41   & 0.017                                            & 0.407          & \cellcolor[HTML]{FFCCC9}0.05         & 0.39       & \cellcolor[HTML]{FFCCC9}0.133     & 0.411  & 0.056                                            & 0.945          & 0.015                                & 0.975      & 0.03                              & 0.99        & \cellcolor[HTML]{FFCCC9}0.047    
\end{tabular}}
\end{table}

\begin{RQ}
{\bf RQ3: Are   some fairness metrics more sensitive  to change than others?} 
\end{RQ}

An ideal metric is responsive to the dataset it examines. An ``insensitive'' metric is one that delivers the same conclusions, no matter what data is being examined. An ``insensitive'' cluster is one containing mostly insensitive metrics. Such insensitive clusters could be ignored since they are  not informative.
 
We measure sensitivity by looking at the variability of our metrics scores using the intra-quartile range  (IQR=$Q_{3} - Q{1}$). For each data set, we found the IQR across all clusters. Next, we \colorbox{mypink}{highlight} the sensitive results; i.e. those with an IQR greater that  $d$*standard deviation. The remaining, unhighlighted results are the  insensitive metrics.

As to what value of $d$ to use in this analysis, we take the advice of a widely cited paper by Sawilowsky~\cite{sawilowsky2009new} (this 2009 paper has 1100 citations). That paper asserts that ``small'' and ``medium'' effects can be measured using $d=0.2$ and $d=0.5$ (respectively). We will analyze this data by splitting the difference looking for differences larger than $d=(0.5+0.2)/2=0.35$.


Turning now to  Table~\ref{tbl:rq4} and Table~\ref{tbl:rq4a}  we see that most clusters have \colorbox{mypink}{highlight} IQR results. However, in  Table~\ref{tbl:rq4}, we see the clusters formed by metrics C16, C18, C20 (individual fairness) and C17, C18, C21, C22, C23, C24 (between group individual fairness) are insensitive. This, in turn, means that we should  \underline{\em not} criticize a fairness analysis that ignores these metrics.

\begin{table}[]
\caption{This table is similar to Table~\ref{tbl:rq4}, showing the sensitivity of the dataset metrics on (a) Baseline; (b) Reweighing (RW). }
\label{tbl:rq4a}
\resizebox{\textwidth}{!}{%
\begin{tabular}{c|cccccc|cccccc|cccccc}
\rowcolor[HTML]{C0C0C0} 
\cellcolor[HTML]{C0C0C0}                      & \multicolumn{6}{c}{\cellcolor[HTML]{C0C0C0}Compas}                                                                                                      & \multicolumn{6}{c|}{\cellcolor[HTML]{C0C0C0}Health}                                                                                                      & \multicolumn{6}{c|}{\cellcolor[HTML]{C0C0C0}German}                                                                                                      \\
\rowcolor[HTML]{C0C0C0} 
\cellcolor[HTML]{C0C0C0}                      & \multicolumn{2}{c}{\cellcolor[HTML]{C0C0C0}Baseline} & \multicolumn{2}{c}{\cellcolor[HTML]{C0C0C0}RW} & \multicolumn{2}{c}{\cellcolor[HTML]{C0C0C0}MFC} & \multicolumn{2}{c|}{\cellcolor[HTML]{C0C0C0}Baseline} & \multicolumn{2}{c}{\cellcolor[HTML]{C0C0C0}RW} & \multicolumn{2}{c}{\cellcolor[HTML]{C0C0C0}MFC} & \multicolumn{2}{c|}{\cellcolor[HTML]{C0C0C0}Baseline} & \multicolumn{2}{c}{\cellcolor[HTML]{C0C0C0}RW} & \multicolumn{2}{c}{\cellcolor[HTML]{C0C0C0}MFC} \\
\rowcolor[HTML]{C0C0C0} 
\multirow{-3}{*}{\cellcolor[HTML]{C0C0C0}MID} & Med           & IQR                                  & Med        & IQR                               & Med                    & IQR                    & Med            & IQR                                  & Med        & IQR                               & Med                    & IQR                    & Med            & IQR                                  & Med        & IQR                               & Med                    & IQR                    \\ \hline
D1                                            & 0.568         & \cellcolor[HTML]{FFCCC9}0.021        & 0.568      & \cellcolor[HTML]{FFCCC9}0.021     & -                      & -                      & 0.804          & 0.02                                 & 0.804      & \cellcolor[HTML]{F8E1E0}0.02      & -                      & -                      & 0.632          & 0.008                                & 0.632      & \cellcolor[HTML]{FFCCC9}0.008     & -                      & -                      \\ \hline
D2                                            & 0.252         & \cellcolor[HTML]{FFCCC9}0.043        & 0          & 0                                 & -                      & -                      & 0.868          & \cellcolor[HTML]{FFCCC9}0.322        & 0.001      & 0                                 & -                      & -                      & 0.298          & \cellcolor[HTML]{FFCCC9}0.105        & 0.002      & 0                                 & -                      & -                      \\ \hline
D3                                            & -0.105        & 0.016                                & 0          & 0                                 & -                      & -                      & -0.313         & \cellcolor[HTML]{FFCCC9}0.067        & 0          & 0                                 & -                      & -                      & -0.097         & \cellcolor[HTML]{FFCCC9}0.033        & 0          & 0                                 & -                      & -                      \\
D4                                            & 0.777         & \cellcolor[HTML]{FFCCC9}0.033        & 1          & 0                                 & -                      & -                      & 0.411          & \cellcolor[HTML]{FFCCC9}0.137        & 1          & 0                                 & -                      & -                      & 0.865          & \cellcolor[HTML]{FFCCC9}0.043        & 1          & 0                                 & -                      & -                     
\end{tabular}}
\end{table}

\begin{table}[!b]
\caption{This table shows the number of classification metrics that move towards or away from the ideal value when either Reweighing or Meta Fair Classifier is used to remove bias in the models. Here ``UF'' shows the number of metrics that moved towards the ideal metric value, while ``FU'' shows the opposite. Finally ``NC''  shows the number of metrics that did not change at all.}
\label{RQ3_answer}\small
\begin{tabular}{l|ccc|ccc}
\rowcolor[HTML]{C0C0C0} 
\cellcolor[HTML]{C0C0C0}                          & \multicolumn{3}{c|}{\cellcolor[HTML]{C0C0C0}\begin{tabular}[c]{@{}c@{}}Reweighing\\ (RW)\end{tabular}} & \multicolumn{3}{c|}{\cellcolor[HTML]{C0C0C0}\begin{tabular}[c]{@{f}c@{}}Meta Fair\\ Classifier (MFC)\end{tabular}} \\
\rowcolor[HTML]{C0C0C0} 
\multirow{-2}{*}{\cellcolor[HTML]{C0C0C0}Dataset} & UF                               & FU                               & NC                               & UF                                   & FU                                   & NC                                  \\ \hline
Adult                                             & 13                               & 13                                & 0                                & 11                                   & 15                                   & 0                                   \\
Compas                                            & 15                               & 7                                & 4                                & 16                                   & 6                                   & 4                                   \\
Health                                            & 17                               & 5                                & 4                                & 17                                   & 7                                    & 1                                   \\
German                                            & 19                               & 6                                & 1                                & 19                                   & 7                                   & 0                                   \\
bank                                              & 16                               & 6                                & 4                                & 15                                   & 7                                   & 4                                   \\
Titanic                                           & 11                               & 15                               & 0                                & 17                                   & 9                                    & 0                                   \\
Student                                           & 15                               & 7                                & 4                                & 12                                   & 10                                   & 4                                  
\end{tabular}
\end{table}

\begin{RQ}
{\bf RQ4: Can we achieve fairness based on all the metrics at the same time?} 
\end{RQ}

Different fairness metrics measure different kinds of bias. If any of the metrics complain about the fairness of the test results, then we can not trust the model blindly, and it should go through further scrutiny and improvement. Bias mitigation algorithms try to make unfair models fairer. Here we are verifying even after applying bias mitigation algorithms; can we achieve fairness based on all the metrics or not? We have chosen two highly cited algorithms from IBM AIF360: Reweighing (RW) by Kamiran et al.~\cite{Kamiran2012} and Meta Fair Classifier (MFR) by Celis et al~\cite{celis2020classification}. 

Table~\ref{RQ3_answer} shows those results collected for seven datasets after using RW and MFC algorithms. For every dataset (row-wise), we  show the number of metrics changed towards or away from its ideal value. In that  table:
   
\begin{itemize}
   \item FU denotes the metrics that changed towards ideal value;
   \item UF denotes the metrics that moved away from the ideal value,
   \item NC means the metrics which did not change.
\end{itemize}
   
Note that  majority of the metrics move towards ``fair'', but there are some metrics that move towards ``unfair''. For Reweighing, some metrics show ``no change'', but we have verified they always remain in the  \textit{fair range}.
   
The main takeaway here is no longer necessary (or even possible) to satisfy all these fairness metrics. While our analysis can reduce dozens of metrics down to ten, there  will still be issues of how to trade-off within this reduced set. Even after applying bias mitigation approaches, some metrics still conflict with others. This finding is similar to the claim made by others:

\begin{itemize}
    \item  Berk et al.~\cite{berk2017fairness} offer an ``Impossibility Theorem'' that says there  is no way to satisfy all kinds of fairness together. 
    \item As Yuriy Brun said at his keynote at \mbox{ICSSP'2020}  ``\textit{we need to work the system in a biased way sometimes}''~\cite{Brun_Keynote}. 
\end{itemize}

\section{Discussion}
We have described all of our results. Here we are summarizing the results in a comprehensible way to reach a stable conclusion. The main idea of this work is to reduce the complexity of measuring fairness. That said, it is imperative we narrate our conclusions in a very easy way. We discuss here three major concerns that arise from \S\ref{sec:results} and try to simplify fairness measurement to our best.

\subsection{Why Not Group Metrics via their Analytical Structure?}
\label{sec:cluster_analysis}
This paper has offered an empirical analysis that many of the metrics in Table~\ref{26_classification_metrics} are synonymous since, when clustered, they fell together into just a few similar groups. In this section, we check if the same conclusions can be achieved from a more analytical analysis that looked at the structure of the equations for the fairness metrics.

Sometimes, a group generated by formula's analytical structure is similar to the clusters we generated above.
For example:
\begin{itemize}
    \item In cluster three (from Table~\ref{26_classification_metrics}), all metrics are based on {\em FDR}, which suggests that both from an empirical and analytical point of view, they should be  similar. 
    \item Also, In  cluster zero, we see that all those  metrics are based on {\em FOR} and error rate.  Intuitively, this seems sensible since   here metrics    try to measure amount of misclassification.
\end{itemize}

That said, as shown by the following three examples, there are  many examples where an equation's analytical structure does {\em not} predict for its empirical cluster.

\begin{itemize}
    \item \textbf{{\em EXAMPLE \#1:}} If we look at cluster five, all six metrics inside this cluster are related to ``between group individual fairness''. This metric is based on the same benefit function: 
        \begin{equation}
        \label{hat}
            y = \hat{y} - y + 1 
        \end{equation}
        
        (For more details on that.  see Table~\ref{tbl:fairness_def} metric id C16). We note that   cluster two is also based on Equation~\ref{hat}, but the metrics inside this cluster represent individual fairness for each group separately. That means 
        
        \begin{quote}
            \textit{Although all metrics inside cluster two and cluster five are based on the same benefit function, they measure different definitions of fairness}.
        \end{quote}
        
        That is, a formal analysis of the analysis might combine these clusters, whereas a data-oriented empirical analysis would argue for their separation.

    \item \textbf{{\em EXAMPLE \#2:}} In cluster four from Table~\ref{tbl:fairness_def}, the metrics C0, C1, C2, C5, C6 and C9 dependent on {\em TPR}, {\em FPR} and {\em FNR}. Recall that  {\em FPR} and {\em FNR} report type one and type two errors ( misclassification on fairness); Now {\em TPR} can be expressed as {\em 1 - FPR}, which means the change in {\em TPR} will mirror changes in {\em FPR}. In contrast, in this cluster, the other two metrics C14 and C15 are based on selection rate (ratio of number of predicted positive and number of instances). Although there is not much similarity in the formula between these two and other metrics in this cluster, we can see they perform similarly when measuring fairness. That is:

    \begin{quote} 
        {\em An analytical analysis does not always reflect the measurement of fairness in the real world scenario.}
    \end{quote}

    Verma et al.~\cite{8452913} in their paper notice a similar phenomenon where they observe that:
    {\em Equal Predictive parity (a measure they explore) should also have equal {\em FDR} ... but when  measured from an empirical point of view, they showed they are not the same.}

    \item \textbf{{\em EXAMPLE \#3:}} In cluster one, metrics C10 and C25 have very different mathematical formulas. C10 is based on {\em FPR} while C25 is based on smoothed EDF-- the Empirical differential fairness. {\em EDF} is calculated based on Dirichlet smoothed base rates for each intersecting group in the dataset, which is based on count of predicted positive. Here as well, we see that

    \begin{quote}
        {\em Two formulas with a different analytical structure can have a similar performance w.r.t. fairness.}
    \end{quote}
\end{itemize}

To summarize the above, we quote Alfred Korzybski, who warned:

\begin{quote}
\centering
    {\em A map is not the territory.}
\end{quote}

While the analytical structure of the formula offers intuitions about the nature of fairness, those intuitions had better be checked via empirical analysis.


\subsection{Is our Empirical Analysis Useful?}
\label{sec:discussion}

We have established the requirement of empirical analysis and we have also done that analysis. We need to find out whether this analysis would be helpful in real-life applications or not. Here we describe various scenarios of fairness contradiction and how our study helps to remove that.

Imagine a college admission decision scenario, where the system might be seen as biased against group~B if applicants from group~A are accepted more than group~B. Here group~A and group~B are divided based on different values of a protected attribute. The college applies a bias mitigation approach to solve this problem using a group fairness metric by changing group A's or B's scoring threshold. Now, if a member of group~A is rejected, while a member of group B has been accepted with an equal or lower score, then the system might be seen as biased against that individual. The main takeaway from this story is that there is a conflict between ``individual fairness'' and ``group fairness''~\cite{binns2019apparent}. 

The concept of fairness is very much application-specific and choosing the appropriate metric is the sole responsibility of the policymaker. An ideal scenario will be building a machine learning model which does not show any kind of bias. However, that is too good to be true. Brun et al. found out that if a model is adjusted to be fair based on one protected attribute (e.g., sex), in some cases model becomes more biased based on another protected attribute (e.g., race)~\cite{FAIRWARE}. Kleinberg and other researchers argue that different notions of fairness are incompatible with each other and hence it is impossible to satisfy all kinds of fairness simultaneously~\cite{kleinberg2016inherent}. Here one thing to remember while doing prediction is that fairness is not the only concern. Prediction performance is the most important goal. Berk et al. found out that accuracy and fairness are competing goals~\cite{Berk}. This trade-off makes the job even more complicated since damaging  model performance   while making it fair may be unacceptable.

As researchers, we know that satisfying all kinds of fairness together is not possible. A policymaker has to choose which fairness definitions are most important for the particular domain and ignore the rest. Our work of dividing fairness tries to make the choice easier, as choosing metrics from a group of 10 options is much simpler than choosing from 30 choices. Using our results of Table~\ref{26_classification_metrics} and Table~\ref{tbl:dataset_metrics}, in a specific domain, if group fairness is more important than individual fairness, then cluster four will be given more priority than clusters two and five (Table~\ref{26_classification_metrics}). Once a cluster is given priority, one or two metrics can be chosen to represent the whole cluster. That means our whole work boils down to minimizing the number of metrics to look at and covering a wide range of fairness. We believe future researchers and industry practitioners will use our work as a guide and that will be the fulfillment of this study.

\subsection{What to do when the metrics contradict each other?}
\label{what_to_do}


We have seen that there are scenarios where fairness metrics contradict each other. According to some metrics, the prediction is fair, where some other metrics disagree. Fairness metrics find out how critical the errors of a prediction model are. It is the decision of the policymaker or the domain expert to choose appropriate fairness metrics based on what kind of bias is more important for the specific domain. For example, consider the following two scenarios:

\begin{itemize}
    \item Suppose we  are predicting if a patient has cancer or not, depending on the symptoms. Here predicting a benign case as malignant is not very dangerous but predicting a malignant case as benign is extremely dangerous. A wrong diagnosis for an actual cancer patient will delay the treatment, and the patient may die. That means \textit{false negative} is more important here. 
    \item Suppose we  are predicting if future performance of a student based on previous records. Here if we predict a good student as bad, that is not that fatal. However, if a student who really needs special attention and help from teachers, is given a good rating then it will be misery for that student. That means \textit{false positive} is more important here. 
\end{itemize}
If we know which metrics look at what kind of error, it will be easier for the decision-maker to choose. That said, based on the guidance we have provided, in case of contradiction among metrics, one metric over another will be given priority.

\section{Threats to Validity \& Future Work}
\label{Limitations}
This paper explores  machine learning methods for software engineering. One issue with any paper like this is a few selection and evaluation biases along with construct and external validity based on the choice of models, datasets, and methods. In the future, we plan to address the apparent threats to validity that this paper has not fully addressed.

\textbf{Construct Validity:} Here, we have used popular \textit{hierarchical} clustering called \textit{agglomerative} approach, as the number of clusters were not known beforehand. In future, we need to experiment with other clustering techniques to check for conclusion stability. This analysis used logistic regression (LR), as much prior work on fairness has also used LR~\cite{IBM, Chakraborty_2020}. Nevertheless, in  future work, we need to  explore some other classification models including DL models. Also, the metric clusters found in Table~\ref{26_classification_metrics} and Table~\ref{tbl:dataset_metrics} are created using the results of our choice ML models, dissimilarity measures, and cutting point in the dendrogram. Thus, choosing one metric from each cluster may contain some risk, and researchers need to be careful while making informed choices about metric selection.

\textbf{Evaluation Bias:} We have used 30 metrics taken from IBM AIF360~\cite{IBM}. We have also covered most of the metrics from Fairkit-learn~\cite{johnson2020fairkit} and Fairlearn~\cite{Fairlearn}. There are other metrics and definitions of fairness, thus the results of this study may not generalize to all available  metrics. But the 30 metrics covered in this study are widely used in the fairness domain~\cite{kearns2018preventing, friedler2019comparative, zafar2017fairness, biswas2020machine, cotter2019optimization}.

\textbf{External Validity:} We have used seven datasets. In the fairness domain, one big challenge is the availability of adequate datasets. It would be insightful to re-run this study on new datasets and also on other domains.
    
\textbf{Sampling Bias:} In this work we used thresholds recommended by IBM AIF360 (``fair'' means  -0.1,0.1 or 0.8,1.2 for different kinds of metric). Future work should explore the sensitivity of our conclusions to changes in those thresholds.

Another issue with sampling bias is that our analysis is based on the data of
Table~\ref{datasets}. We recommend that when new data becomes available, we test the conclusions of this paper against that new data. 
That would not be
an arduous task (and to simplify that task, we  have placed all our scripts online in order).


\section{Conclusion}
\label{Conclusion}
Fairness is a rapidly evolving domain and the number of fairness metrics is increasing exponentially. While performing our literature review we saw the current practice in this domain is to relying on a handful of metrics and ignoring the rest. But which metrics can be ignored? Which are essential?

To answer these questions, this paper has experimented with the  following {\em metrics selection tactic}:
When applied, the paper reported that this tactic could reduce dozens of metrics to just a handful. We found:
\begin{itemize}
    \item RQ1 showed that all the metrics do not agree with each other when labeling a model as fair or unfair.
    \item RQ2 showed that metrics can be clustered together based on how they measure bias. Each of the resultant clusters measures different types of bias and selecting one metric from each cluster should be representative enough to measure increase or decrease in bias in other metrics in the same cluster.
    \item RQ3 showed that we could ignore at least two of those clusters, since they were not ``sensitive''. Recall that by ``insensitive'' clusters, we mean those where changes to the data did not change the fairness scores.
    \item RQ4 showed this reduced set actually predicts for different things. That said, it is no longer necessary (or even possible) to satisfy all these fairness metrics. 
\end{itemize}
From these results, we argue that:
\begin{itemize}
\item There are  many  spurious fairness
metrics; i.e. metrics that   measure very similar things.
    \item To  simplify   fairness testing,  just (a)~determine what type of fairness is desirable (for a list of types, see Table~\ref{26_classification_metrics} and Table~\ref{tbl:dataset_metrics} ); then (b) look up those types in our clusters; then (c)~just test for one item per cluster.
    \item   While this approach does not completely remove all issues with   fairness testing, it does reduce a very complex problem of (say) 30 metrics to a much smaller and manageable set.
\item
Also, the methods of this paper could be used as a litmus test to prune away  spurious new  metrics that merely report the same thing as existing metrics.
\end{itemize}

\section{Acknowledgement}
The work was partially funded by LAS and NSF grant \#1908762.


\bibliographystyle{ACM-Reference-Format}
\bibliography{main}






\end{document}